\documentclass{article}
\usepackage{amssymb}

\usepackage[square,numbers,sort&compress]{natbib}

\usepackage{bibentry}

\usepackage{amsmath}

\usepackage{amssymb}

\usepackage{mathtools}

\usepackage{xspace}

\usepackage{cancel}

\usepackage{microtype}

\usepackage{algorithm}
\usepackage[noend]{algpseudocode}

\usepackage[inline]{enumitem}

\usepackage{multicol}

\usepackage{booktabs}

\usepackage{subfig}

\usepackage[usenames,dvipsnames,table]{xcolor}

\usepackage[most]{tcolorbox}

\usepackage{nag}

\usepackage{todonotes}
\setuptodonotes{
  backgroundcolor=pink!30,
  textcolor=red!90!black,
  linecolor=red!90!black,
  bordercolor=red!90!black,
  caption={},  
}

\usepackage{hyperref}

\definecolor{deeppink}{rgb}{1.0, 0.08, 0.58}
\hypersetup{
  colorlinks = true,  
  urlcolor   = deeppink,  
  linkcolor  = blue,  
  citecolor  = red,  
  pdfview    = XYZ,  
}

\usepackage[nameinlink]{cleveref}  

\AtBeginEnvironment{appendices}{\crefalias{section}{appendix}}
\AtBeginEnvironment{appendices}{\crefalias{subsection}{appendix}}
\AtBeginEnvironment{appendices}{\crefalias{subsubsection}{appendix}}

\crefname{equation}{Eq.}{Eqs.}
\crefname{algorithm}{Alg.}{Algs.}
\crefname{figure}{Fig.}{Figs.}
\crefname{table}{Table}{Tables}
\crefname{tabular}{Table}{Tables}
\crefname{section}{Sec.}{Sections}
\crefname{appendix}{App.}{Appendices}

\creflabelformat{equation}{#2\textup{#1}#3}

\allowdisplaybreaks

\usepackage{hyperxmp}

\usepackage[acronym,toc]{glossaries}

\usepackage[export]{adjustbox}

\usepackage{tikz}

\usepackage[bottom,flushmargin]{footmisc}
\makeatletter  
\def\blfootnote{\gdef\@thefnmark{}\@footnotetext}
\makeatother

\usepackage{lineno}

\usepackage{dblfloatfix}

\usepackage{lipsum}

\tcbuselibrary{theorems}
\newtcbtheorem{boxtheorem}{\it Theorem}{colback=gray!30, colframe=black!60, title=\textit}{thm}

\usepackage{gensymb}

\usepackage{graphicx}

\usepackage{multirow}

\usepackage{makecell}

\newcolumntype{?}{!{\vrule width 1pt}}

\usepackage{pifont}

\usepackage{color, soul}

\usepackage{appendix}

\newsavebox\CBox

\definecolor{darkred}{rgb}{0.55, 0.0, 0.0}

\newcommand{\ie}{\textit{i.e.}\xspace}
\newcommand{\eg}{\textit{e.g.}\xspace}

\newcommand{\A}{{\mathcal{A}}}
\newcommand{\TA}{T\hspace{-0.2em}\A}

\renewcommand{\H}{{\mathcal{H}}}
\newcommand{\aprev}{a_\mathrm{prev}}

\newcommand{\lcfm}{\mathcal{L}_\mathrm{CFM}}
\newcommand{\Tpred}{T_\mathrm{pred}}
\newcommand{\Tchunk}{T_\mathrm{chunk}}

\renewcommand{\mid}{\,\vert\,}
\newcommand{\bigmid}{\,\big\vert\,}
\newcommand{\N}{\mathcal{N}}

\newcommand{\dataset}{\mathcal{D}}

\newcommand{\weights}{\theta}

\newcommand{\dd}{\mathrm{d}}

\newcommand{\R}{\mathbb{R}}
\newcommand{\Rp}{\R_{\geq 0}}

\newcommand{\EBigSubscript}[3]{\mathbb{E}\raisebox{#1ex}{\fontsize{#2pt}{#2pt}\selectfont\hspace{0.2ex}$_{#3}$}}

\newcommand{\redcolor}{red!10}
\newcommand{\greencolor}{green!10}
\newcommand{\bluecolor}{cyan!10}

\newcommand{\redbox}[1]{\raisebox{0.5ex}{\fcolorbox{black}{\redcolor}{\rule{0pt}{#1}\rule{#1}{0pt}}}\xspace}
\newcommand{\greenbox}[1]{\raisebox{0.5ex}{\fcolorbox{black}{\greencolor}{\rule{0pt}{#1}\rule{#1}{0pt}}}\xspace}
\newcommand{\bluebox}[1]{\raisebox{0.5ex}{\fcolorbox{black}{\bluecolor}{\rule{0pt}{#1}\rule{#1}{0pt}}}\xspace}

\usepackage[final]{corl_2025} 

\newcommand{\website}{https://streaming-flow-policy.github.io}

\title{
	Streaming Flow Policy\\
  \vspace{0.1em}
	{
		\fontsize{15}{15}\normalfont
        Simplifying diffusion$/$flow-matching policies by\\
        \vspace*{-0.2em}
        treating action trajectories as flow trajectories
	}
  \\
	\vspace{0.2em}
	{
		\normalfont
		\normalsize
		{\small Website:} \href{\website}{\color{deeppink}\texttt{\website}}
	}
  \\  
  \vspace{0em}
}

\newcommand{\add}[1]{\begingroup\ignorespaces#1\endgroup}

\newcommand{\authorspace}{\;\;}
\newcommand{\authorcomma}{{{\normalfont, }}}
\author{
  \textbf{Sunshine Jiang}\authorcomma \authorspace
  \textbf{Xiaolin Fang}\authorcomma \authorspace
  \textbf{Nicholas Roy}\authorcomma\\
  \textbf{Tomás Lozano-Pérez}\authorcomma \authorspace
  \textbf{Leslie Kaelbling}\authorcomma \authorspace
  \textbf{Siddharth Ancha}
  \\
  Massachusetts Institute of Technology
}

\begin{document}
\maketitle


\vspace*{-1.3em}
\begin{abstract}  
    Recent advances in diffusion$/$flow-matching policies have enabled imitation learning of complex, multi-modal action trajectories.
    However, they are computationally expensive because they sample a \textit{trajectory of trajectories}—a diffusion$/$flow trajectory of action trajectories.  
    They discard intermediate action trajectories, and must wait for the sampling process to complete before any actions can be executed on the robot.  
    We simplify diffusion$/$flow policies by \textit{treating action trajectories as flow trajectories}.  
    Instead of starting from pure noise, our algorithm samples from a narrow Gaussian around the last action.
    Then, it incrementally integrates a velocity field learned via flow matching to produce a sequence of actions that constitute a \textit{single} trajectory.  
    This enables actions to be streamed to the robot on-the-fly \textit{during} the flow sampling process,
    and is well-suited for receding horizon policy execution.
    Despite streaming, our method retains the ability to model multi-modal behavior.  
    We train flows that \textit{stabilize} around demonstration trajectories to reduce distribution shift and improve imitation learning performance.  
    Streaming flow policy outperforms
    prior methods
    while enabling faster policy execution and tighter sensorimotor loops for learning-based robot control.
\end{abstract}

\vspace*{1.1em}
{
\centerline
{
  \includegraphics[trim=0 0 0 0,clip,width=0.65\textwidth]{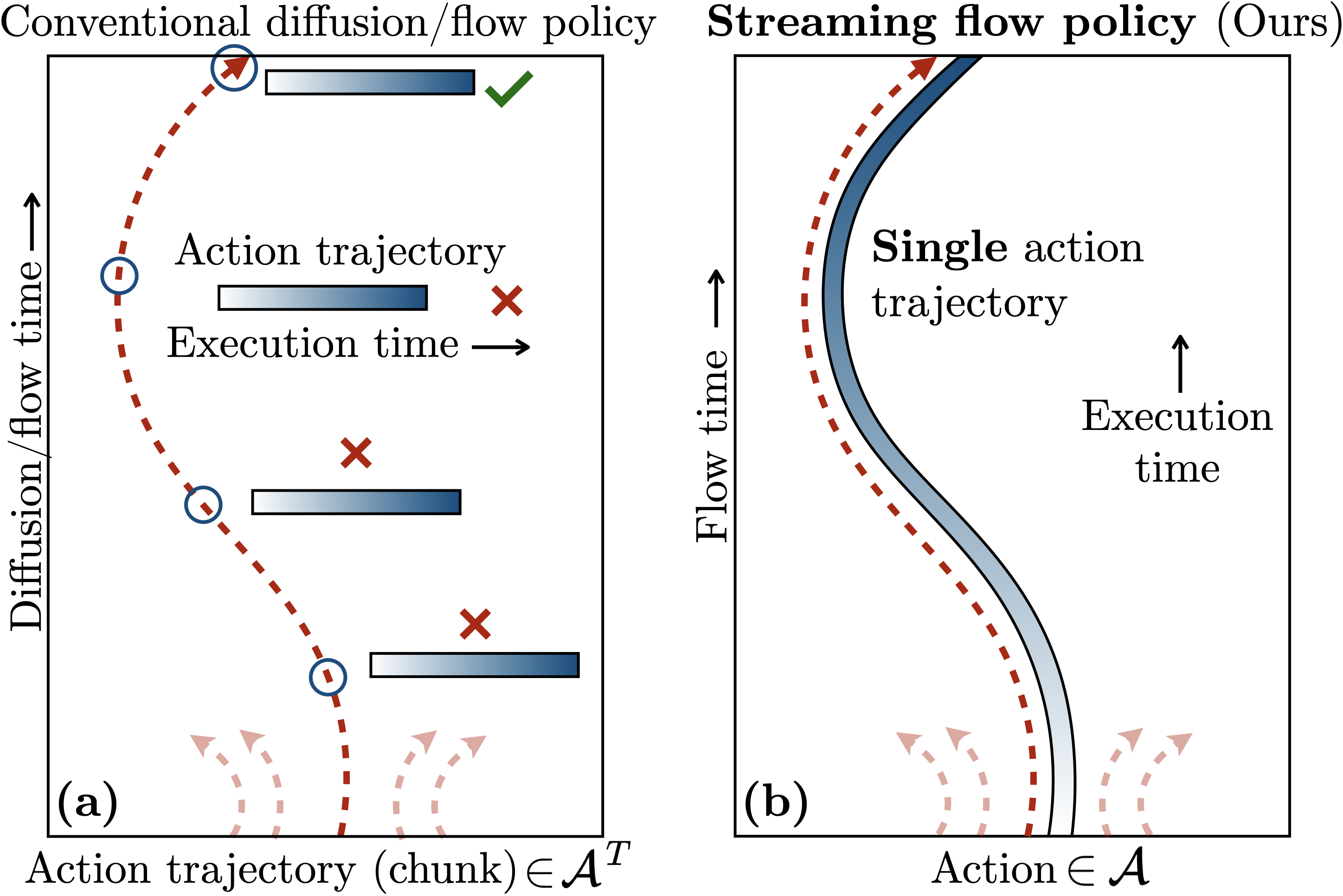}
}
\vspace*{-0.4em}
\captionof{figure}
{
  \small
  \textbf{(a)}
  Diffusion policy~\cite{chi2023diffusion} and flow-matching policy~\cite{black2410pi0} input a history of observations (not shown) to predict a ``chunk'' of future robot actions.
  The $x$-axis represents the action space, and the +$y$-axis represents increasing diffusion$/$flow timesteps.
  Conventional diffusion$/$flow policies sample a ``trajectory of trajectories'' --- a diffusion$/$flow trajectory of action trajectories.
  They discard intermediate trajectories,
  and must wait for the diffusion$/$flow process to complete before the first actions can be executed on the robot.
  \textbf{(b)}
  We simplify diffusion$/$flow policies by \textit{treating action trajectories as flow trajectories}.
  Our flow-matching algorithm operates in {action space}. Starting from a noised version of the last executed action, it incrementally generates a sequence of actions that constitutes a \textit{single} trajectory.
  This aligns the ``time'' of the flow sampling process with the ``execution time'' of the action trajectory.
  Importantly, actions can be streamed to the robot's controller on the fly \textit{during} the flow sampling process, while retaining the ability to model multi-modal trajectory distributions.
}
\label{fig:pull-figure}
\vspace*{0.5em}
}

\clearpage  


{
\begin{figure}[t!]
\centerline
{
  \includegraphics[trim=0 0 0 0,clip,width=1.15\textwidth]{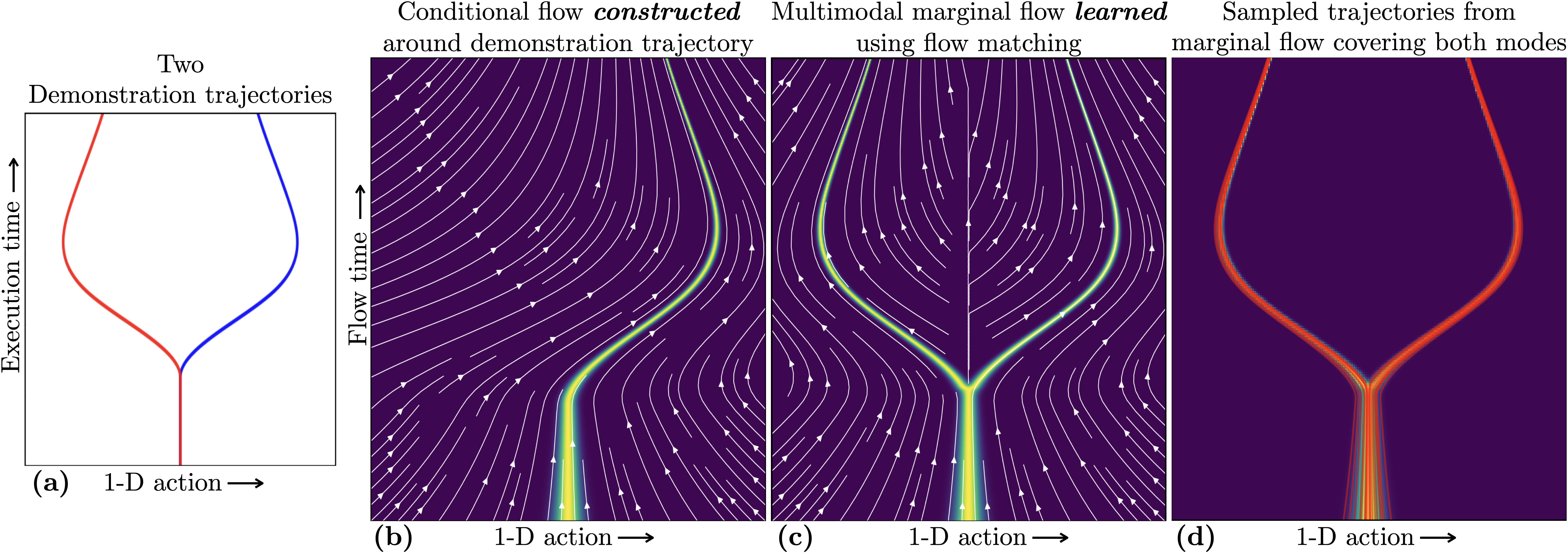}
}
\vspace*{-0.5em}
\captionof{figure}
{
  \small
  \textbf{(a)}
  To illustrate our method, we consider a toy example of 1-D robot actions with two demonstration trajectories shown in blue and red.
  \textbf{(b)}
  Given a demonstration trajectory sampled from the training set (\eg the blue one), we first analytically \textit{construct} a conditional flow \ie an initial action distribution and a velocity field.
  The constructed flow samples trajectories from a thin Gaussian tube around the demonstration trajectory.
  Using the constructed velocity field as targets, we \textit{learn} a marginal velocity field via flow matching~\cite{lipman2023flow}, shown in \textbf{(c)}. 
  The learned velocity field has the property that its induced marginal distribution over actions at each horizontal time slice matches the training distribution.
  \textbf{(d)}
  The initial action at $t=0$ is sampled from a narrow Gaussian centered at the most recently executed action.
  Then, we iteratively integrate the learned velocity field to generate an action trajectory.
  Sampled trajectories (shown in red) cover both behavior modes in the training data.
  \textbf{(b)} We find that constructing conditional flows that \textit{stabilize} around demonstration trajectories reduces distribution shift and improves imitation learning performance.
  The main takeaway is that our method is able to both represent multi-modal distributions over action trajectories
  like diffusion$/$flow policies, while also iteratively generating actions that can be streamed during the flow sampling process, enabling fast and reactive policy execution.
}
\label{fig:1d-example}
\end{figure}
}
\section{Introduction}

Recent advances in robotic imitation learning, such as diffusion policy~\cite{chi2023diffusion,team2024octo} and flow-matching policy~\cite{black2410pi0,zhang2024affordance,ye2024efficient} have enabled robots to learn complex, multi-modal action distributions for challenging real-world tasks such as cooking, laundry folding, robot assembly and navigation~\cite{sridhar2024nomad}.
They take a history of observations as input, and output a sequence of actions (also called an ``action chunk").
Conventional diffusion$/$flow policies represent a direct application of diffusion models~\cite{sohl2015deep,ho2020denoising} and flow-matching~\cite{lipman2023flow} to robot action sequences ---
they formulate the generative process as probabilistic transport in the space of action \textit{sequences}, starting from pure Gaussian noise.
Therefore, diffusion$/$flow policies represent a ``\textit{trajectory of trajectories}'' --- a diffusion$/$flow trajectory of action trajectories (\cref{fig:pull-figure}a).
This approach has several drawbacks.
The sampling process discards all intermediate action trajectories, making diffusion$/$flow policies computationally inefficient.
Importantly, the robot must wait for the diffusion$/$flow process to complete before executing any actions.
Thus, diffusion$/$flow policies often require careful hyper-parameter tunning to admit tight control loops.

In this work, we propose a novel imitation learning framework that harnesses the temporal structure of action trajectories.
We simplify diffusion$/$flow policies by treating \textit{action trajectories as flow trajectories} (\cref{fig:pull-figure}b).
Our aim is to learn a flow transport
in the \textit{action space} $\A$, as opposed to trajectory space $\A^T$.
Unlike diffusion$/$flow policies that start the sampling process from pure Gaussian noise (in $\A^T$),
our initial sample comes from a \textit{narrow} Gaussian centered around the most recently generated action (in $\A$).
Then, we iteratively integrate a learned velocity field to generate a sequence of future actions that forms a \textit{single} trajectory.
The ``flow time" --- indicating progress of the flow process ---
coincides with execution time of the sampled trajectory.
Iteratively generating the sequence of actions allows the actions to be \textit{streamed} to the robot's controller on-the-fly \textit{during} the flow generation process,
significantly improving the policy's speed and reactivity.

We show how a streaming flow policy with the above desiderata can be learned using
flow matching~\cite{lipman2023flow}.
Given an action trajectory from the training set (\cref{fig:1d-example}a), we \textit{construct} a velocity field conditioned on this example that samples paths in a narrow Gaussian ``tube" around the demonstration (\cref{fig:1d-example}b).

Our training procedure is remarkably simple --- we regress a neural network $v_\weights(a, t \mid h)$ that takes as input (i) an observation history $h$, (ii) flow timestep $t \in [0, 1]$, and (iii) action $a$, to match the constructed velocity field.
We are able to re-use existing architectures for diffusion$/$flow policy while only modifying the input and output dimension of the network from $\A^T$ to $\A$.
Flow matching guarantees that the \textit{marginal} flow learned over all training trajectories, as shown in \cref{fig:1d-example}(c, d), is multi-modal.
Specifically, the marginal distribution of actions at \textit{each timestep} $t$ matches that of the training distribution.
Our approach thus retains diffusion$/$flow policy's ability to represent multi-modal trajectories while allowing for streaming trajectory generation.

How should we construct the target velocity field?
Prior work~\cite{block2023provable} has shown that low-level stabilizing controllers can reduce distribution shift and improve theoretical imitation learning guarantees.
We leverage the flexibility of the flow matching framework to construct velocity fields that \textit{stabilize} around a given demonstration trajectory,
by adding velocity components that guide the flow back to the demonstration.
In our experiments, we find that stabilizing flow significantly improves performance.

Our method can leverage two key properties specific to robotics applications: (i) robot actions are often represented as position setpoints of the robot's joints or end-effector pose that are tracked by a low-level controller, (ii) the robot's joint positions$/$end-effector poses can be accurately measured via proprioceptive sensors (\eg joint encoders) and forward kinematics.
Streaming flow policy can not only imitate action trajectories, but is especially suited to imitate \textit{state trajectories} when a stiff controller is available that can closely track state trajectories.
In this case, the flow sampling process can be initialized from the known ground truth robot state instead of the state predicted from the previous chunk.
This reduces uncertainty and error in the generated trajectory.

Unlike diffusion$/$flow policies, streaming flow policy is only guaranteed to match the marginal distribution of actions at each timestep, but not necessarily the joint distribution.
Consequently, our method can produce trajectories that are compositions of segments of training trajectories, even if the composition was not part of the training dataset.
While this may be seen as a limitation of our method, we argue that for most robotics tasks, compositionality is not only valid, but a desirable property that requires fewer demonstrations.
Furthermore, while streaming flow policy is unable to capture global constraints that can only be represented in the joint distribution,
it can learn local constraints such as joint constraints, and convex velocity
constraints; see \cref{sec:limitations} for more details.
In practice, we find that streaming flow policy performs comparably to diffusion policy while being significantly faster.

 
{
\newcommand{\emphasis}[1]{#1}
\newcommand{\hthickness}{1.1pt}
\newcommand{\vthickness}{1.1pt}
\newcolumntype{I}{!{\vrule width \vthickness}}
\newcolumntype{|}{!{\vrule width 0.8pt}}
\newcolumntype{:}{!{\vrule width 0pt}}

\newcolumntype{M}[1]{>{\centering\arraybackslash}m{#1}}

\renewcommand{\arraystretch}{1.02}  

\begin{table}[t!]
    \small
    \setlength\tabcolsep{6pt}
    \centerline
    {
    \begin{tabular}{IM{1.4cm}|M{9.5cm}|M{1.75cm}I}
     \Xhline{\hthickness}
     \textbf{Symbol} & \textbf{Description} & \textbf{Domain} \\
     \Xhline{\hthickness}
     $\Tpred$ & Prediction time horizon of trajectories during training& $\R^+$ \\
     \hline
     $\Tchunk$ & Time horizon of action chunk during inference& $\R^+$ \\
     \hline
     $t$ & Flow time $=$ execution time rescaled from $[0, \Tpred]$ to $[0, 1]$ & $[0, 1]$ \\
     \hline
     $a$ & Robot action (often a robot configuration) & $\A$ \\
     \hline
     $v$ & Action velocity & $\TA$ \\
     \hline
     $o, h$ & Observation, Observation history & $\mathcal{O}, \H$ \\
     \hline
     $\xi$ & Action trajectory (chunk), where time is rescaled from $[0, \Tpred]$ to $[0, 1]$ & $[0,1] \to \A$ \\
     \hline
     \rule{0pt}{2.5ex}$\dot{\xi}$  & Time derivative of action trajectory: $\dot{\xi}(t) = \frac{d}{dt}\xi(t)$ & $[0,1] \to \TA$ \\
     \Xhline{\hthickness}
     $p_\dataset(h, \xi)$ & 
     \makecell{
        Distribution of observation histories and future action chunks.\\
        Training set is assumed to be sampled from this distribution.
    }
     & $\Delta(\H \times [0, 1] \to \A)$ \\
     \hline
     $v_\weights(a, t \mid h)$ & Learned marginal velocity field with network parameters $\weights$ & $\TA$ \\
     \hline
     $\hspace{-0.2em}v_\xi(a, t)$ & Conditional velocity field for demonstration $\xi$ & $\TA$ \\
     \hline
     $\hspace{-0.2em}p_\xi(a \mid t)$ & Marginal probability distribution over $a$ at time $t$ induced by $v_\xi$ & $\Delta(\A)$ \\
     \hline
     $v^*(a, t \mid h)$ & Optimal marginal velocity field under data distribution $p_\dataset$ & $\TA$ \\
     \hline
     $\hspace{-0.4em}p^*(a \mid t, h)$ & Marginal probability distribution over $a$ at time $t$ induced by $v^*$ & $\Delta(\A)$ \\
     \Xhline{\hthickness}
     $k, \sigma_0$ & Stabilizing gain, Initial standard deviation & $\Rp$, $\R^+$ \\
     \Xhline{\hthickness}
    \end{tabular}
    }
    \vspace{0.3em}
    \caption{
        \small
        Mathematical notation used throughout the paper.
    }
    \label{table:notation}
    \vspace*{-3em}
\end{table}
}

\section{Background and problem formulation}

We consider the problem of imitating sequences of \textit{future} actions $a \in \A$ from histories of observations $h \in \H$ as input,
where a history $h = \{o_i\}_{i=1}^K$ is a finite sequence of observations $o_i \in \mathcal{O}$.
The time horizon $\Tpred \in \R^+$ of the trajectory to be predicted can be an arbitrary hyperparameter.
For simplicity, we re-scale the time interval to $[0, 1]$ by dividing by $\Tpred$.
Therefore, we represent an action trajectory as \mbox{$\xi : [0, 1] \to \A$}.
We assume an unknown data generating distribution $p_\dataset(h, \xi)$ of inputs and outputs, from which a finite training dataset $\dataset = \{(h_i, \xi_i)\}_{i=1}^N$ of $N$ tuples is sampled.
See \cref{table:notation} for a complete list of notation.
Our aim is to learn a policy that outputs a potentially multi-modal distribution over future trajectories $\xi$ given a history of observations $h$.

\add{
\textbf{Assumptions:} We make two key assumptions: \textit{(i)} the action space $\A$ is continuous (more generally, a smooth manifold), and \textit{(ii)} action trajectories $\xi : [0, 1] \to \A$ are continuous and differentiable with respect to time.
These assumptions are necessary to compute time-derivatives of action trajectories when treating
action trajectories as differentiable flow trajectories.
Most physical action spaces used in robotics, such as joint angles or end-effector poses, indeed satisfy these assumptions.
}

\textbf{Velocity fields:} We formulate \textit{streaming flow policy}, with model parameters $\weights$, as a history-conditioned velocity field $v_\weights\left(a, t \mid h\right)$.
For a given history \mbox{$h \in \H$}, \mbox{$t \in [0, 1]$}, and action \mbox{$a \in \A$}, the model outputs a velocity in the tangent bundle $\TA$ of $\A$.
The velocity field is a neural ordinary differential equation (ODE)~\cite{chen2018neural}.
Given an initial action $a(0)$, the velocity field induces trajectories $a(t)$ in action space by specifying the instantaneous time-derivative of the trajectory ${d a}/{d t} = v_\weights\left(a, t \mid h\right)$.

\textbf{Flows:}
The pairing of $v_\weights\left(a, t \mid h\right)$ with an initial probability distribution over $a(0)$ is called a \textit{continuous normalizing flow}~\cite{chen2018neural,grathwohl2018scalable} (simply referred to as ``\textit{flow}").
A flow transforms the initial action distribution to a new distribution $p_\weights\left(a \mid t, h\right)$, for every $t \in [0, 1]$, in a deterministic and invertible manner.
We want streaming flow policy to start sampling close to the action $\aprev$ that was most recently executed.
This is the final action that was computed in the previous action chunk.
When imitating state trajectories instead of action trajectories, we set $\aprev$ to the current known robot state.
Invertible flows require the initial probability distribution over $a(0)$ to have non-zero probability density on the domain $\A$ to be well-behaved.
Therefore, we chose a narrow Gaussian distribution centered at $\aprev$ with a small variance $\sigma_0^2$.
A trajectory is generated by sampling from the initial distribution and integrating the velocity field as:
\begin{align}
    a(t) = a_0 + \hspace{-0.3em}\int_0^t\hspace{-0.5em} v_\weights\hspace{-0.2em}\left(a(\tau), \tau \bigmid h\right) d\tau \hspace{0.6em} \text{where} \hspace{0.6em} a_0 \sim \N\hspace{-0.2em}\left(\aprev, \sigma_0^2\right)
    \label{eq:flow-integration}
\end{align}
Importantly, standard ODE solvers can perform forward finite-difference integration auto-regressively, where integration at time $t$ depends only on previously computed actions $a(\tau), \tau \leq t$.
This property allows us to stream actions \textit{during} the integration process, without needing to wait for the full trajectory to be computed.
Next, we describe how we analytically construct conditional velocity fields given a trajectory $\xi$.
Then, we will use them as targets to learn $v_\weights\left(a, t \mid h\right)$ using flow matching~\cite{lipman2023flow}.

\section{Analytically constructing conditional velocity fields}
\label{sec:conditional-velocity-field}

Given an action trajectory $\xi$, we first analytically construct a stabilizing \textit{conditional} flow that travels closely along $\xi$.
This will be used as a target to train a neural network velocity field.
In particular, we construct a velocity field $v_\xi(a, t)$ and an initial distribution $p^0_\xi(a)$ such that the induced marginal probability distributions $p_\xi(a \mid t)$ form a thin Gaussian ``tube" around $\xi$.
By ``Gaussian tube", we mean that $p_\xi(a \mid t)$ is a narrow Gaussian distribution centered at $\xi(t)$ for every $t \in [0, 1]$.
This is illustrated in \cref{fig:1d-example}(a,b).
We construct the stabilizing conditional flow as:
\begin{align}
    v_\xi(a, t) \ = \underbrace{\ \dot{\xi}(t)\ }_{\hspace{-3.5em}\text{Trajectory velocity}} - \underbrace{k(a - \xi(t))}_{\text{Stabilization term}}
    \quad \text{and} \quad
    p_\xi^0(a) \ =\  \mathcal{N}\left(a ~\vert~ \xi(0) \,,\, \sigma_0^2\right)
    \label{eq:stabilizing-conditional-flow-velocity-field}
\end{align}
The initial distribution $p_\xi^0(a)$ is a narrow Gaussian centered at the initial action $\xi(0)$ with a small standard deviation $\sigma_0$.
The velocity has two components.
The \textit{trajectory velocity} is the time-derivative of the action trajectory $\xi$ at time $t$, and does not depend on $a$.
This term serves to move along the direction of the trajectory.
The \textit{stabilization term} is a negative proportional error feedback that corrects deviations from the trajectory.
Controllers that stabilize around demonstration trajectories are known to reduce distribution shift and improve theoretical imitation learning guarantees~\cite{block2023provable}.
We empirically observe that the stabilizing term produces significantly more robust and performant policies, compared to setting $k = 0$.
We note that our framework leverages time derivatives of action trajectories $\dot{\xi}(t)$ during training, which are easily accessible, in addition to $\xi(t)$.
This is in contrast to conventional diffusion$/$flow policies that only use $\xi(t)$ but not $\dot{\xi}(t)$.
Throughout this paper, the term `velocity' refers to $\dot{\xi}(t)$, and not the physical velocity of the robot.
While they may coincide for certain choices of the action space $\A$, $\dot{\xi}(t)$ may not represent any physical velocity.

\textbf{\textit{Theorem 1:}} The stabilizing conditional flow given by \cref{eq:stabilizing-conditional-flow-velocity-field} induces the following per-timestep marginal distributions over the action space:
\begin{align}
    p_\xi(a \mid t) &= \mathcal{N}\left(a \bigmid \xi(t),\  \sigma_0^2 e^{-2kt}\right)
    \label{eq:stabilizing-conditional-flow-marginal-distribution}
\end{align}
\textit{Proof:} See \cref{app:proof-theorem-1}. The distribution of states sampled at any timestep $t \in [0, 1]$ is a Gaussian centered at the trajectory $\xi(t)$.
Furthermore, the standard deviation $\add{\sigma_t =} \sigma_0e^{-kt}$ starts from $\sigma_0$ at $t=0$ and decays exponentially with time at rate $k$.

\section{Learning objective for velocity fields to match marginal action distributions}
\label{sec:learning-marginal-velocity-fields}

Let $p_\dataset(h, \xi)$ denote the unknown data generating distribution from which the training dataset is sampled.
The conditional velocity field  $v_\xi(a, t)$ defined in \cref{sec:conditional-velocity-field} models a single action trajectory.
If multiple behaviors $\xi$ are valid for the same input history $h$,
how can we learn a velocity field $v(a, t \mid h)$ that represents multi-modal trajectory distributions?
We use flow matching with $v_\xi(a, t)$ the target.
The conditional flow matching loss~\cite{lipman2023flow} for a history-conditioned velocity field $v(a, t \mid h)$ is defined as:
\begin{align}
    \lcfm(v, p_\dataset) =~ 
    \EBigSubscript{0.1}{12}{(h,\,\xi) \sim p_\dataset}~\EBigSubscript{0.1}{12}{t \sim {U}[0,1]}~\EBigSubscript{0.1}{12}{a \sim p_\xi\left(a \mid t\right)} \, \big\|v(a, t \mid h) - v_\xi(a, t)\big\|_2^2
    \label{eq:ideal-cfm-loss}
\end{align}
This is simply an expected $L_2$ loss between a candidate velocity field $v(a, t \mid h)$ and the the analytically constructed conditional velocity field $v_\xi(a, t)$ as target.
The expectation is over histories and trajectories under the probability distribution $p_\dataset(h, \xi)$, time $t$ sampled uniformly from $[0, 1]$, and action $a$ sampled from the constructed conditional flow known in closed-form in \cref{eq:stabilizing-conditional-flow-marginal-distribution}.
The following theorem characterizes the per-timestep marginal distributions induced by the minimizer of this loss:

\textbf{\textit{Theorem 2:}} The minimizer \mbox{$v^* = \arg \min_v \lcfm(v, p_\dataset)$}
induces the following per-timestep marginal distribution for each $t \in [0, 1]$ and observation history $h$:
\begin{align}
    p^*(a \mid t, h)
    = \int_\xi p_\xi(a \mid t)\,p_\dataset(\xi \mid h) \, \dd\xi \,
    \add{ = \int_\xi \N \left( a \mid \xi(t), \sigma_t^2 \right) \,p_\dataset(\xi \mid h) \, \dd\xi \,~\text{where}~\, \sigma_t = \sigma_0 e^{-kt}}
    \label{eq:averaged-marginal-distributions}
\end{align}
\textit{Proof:} This is a direct consequence of the flow matching theorems (Thms. 1 and 2) in~\citet{lipman2023flow}.
Intuitively, the per-timestep marginal distribution induced by the minimizer of $\lcfm$ is the \textit{average} of per-timestep marginal distributions of constructed conditional flows  $p_\xi(a \mid t)$, over the  distribution of future trajectories in $p_\dataset(\xi \mid h)$ that share the same observation history $h$.
\add{Since our conditional flows are constructed to sample narrow Gaussian tubes around demonstrated action sequences, the optimal $v^*$ will produce a \textit{multi-modal} mixture distribution over future actions from the data-generating distribution consistent with observation history $h$, convolved with small noise $\sigma_t$.}

Matching the per-timestep marginal distributions is desirable and necessary for representing multi-modal distributions.
Consider the example in \cref{fig:1d-example} that constructs two conditional flows, one that samples actions to the right ($a > 0$), and the other that samples actions to the left ($a < 0$).
In order for a learned model to sample both modes with probability 0.5 each, its per-timestep marginal distribution must match the averaged per-timestep marginal distributions of conditional flows.
Unlike flow policies~\cite{black2410pi0,zhang2024affordance,ye2024efficient} that only require matching the target distributions at \mbox{$t=1$}, our method leverages the fact that flow matching~\cite{lipman2023flow} matches the marginal distributions
at \textit{all} timesteps $t \in [0, 1]$.

\newcommand{\commentIndent}{1.1em}
\newcommand{\makeColor}{\color{blue}}

\begin{figure}[t!]
    \centerline{
    \begin{minipage}[t]{0.48\textwidth}
        \begin{algorithm}[H]
            \caption{\small Training algorithm}
            \begin{algorithmic}[1]
                {\Require { Training set $\dataset = \{(h_i, \xi_i)\}_{i=1}^N$}}, $\Tpred$
                \Statex \hspace{1.5em} {\raisebox{0.25ex}{\tiny $\bullet$}} {\small\it  $\xi$ has time horizon $\Tpred$ rescaled to $[0, 1]$}
                \While{\text{not converged}}
                    \State $(h, \xi) \sim \dataset$
                    \State $t \sim \mathrm{Uniform}(0, 1)$
                    \State $a \sim p_\xi(a \mid t)$ \quad (defined in \cref{eq:stabilizing-conditional-flow-marginal-distribution})
                    \State $\weights \leftarrow \weights - \hspace{0em} \lambda \nabla_\weights \hspace{-0.2em} \underbrace{\|v_\xi(a, t) - v_\weights(a, t \mid h)\|^2}_{{\text{\fontsize{8}{8}\selectfont Conditional flow matching loss}}}$
                \EndWhile
                \State \Return $v_\weights$
            \end{algorithmic}
            \label{alg:training}
        \end{algorithm}
    \end{minipage}
    \hspace{0.5em}
    \begin{minipage}[t]{0.57\textwidth}
        \begin{algorithm}[H]
            \caption{\small Inference algorithm}
            \begin{algorithmic}[1]
                {\Require $v_\weights(a, t \mid h)$}, $\Tpred$, $\Tchunk$, $\Delta t$
                \State $h, a \leftarrow \{\}~,~ q_\mathrm{curr}$ {(\small current robot configuration)}
                \While{True}
                    \State $t,  h_\mathrm{chunk} \leftarrow 0, h$
                    \State \textbf{if} {\small imitating state}: $a \leftarrow q_\mathrm{curr}$
                    \While{$t ~\leq~ \Tchunk / \Tpred$} {~\small\makeColor {\textit{// open loop}}}
                        \State $o \leftarrow \texttt{Execute}(a)$  {~\small\makeColor {\textit{// stream action during flow}}}
                        \label{line:algo-inference-execute}
                        \State $h \leftarrow h \cup \{o\}$
                        \State $a \leftarrow a + v_\weights(a, t \mid h_\mathrm{chunk}) \Delta t$ {~\small\makeColor {\textit{// integration step}}}
                        \State $t \leftarrow t + \Delta t$
                    \EndWhile
                \EndWhile
            \end{algorithmic}
            \label{alg:inference}
        \end{algorithm}
    \end{minipage}
    }
\end{figure}

\section{Training and inference algorithms for streaming flow policy}
\label{sec:training-inference-algorithms}

\textbf{Training:} While we do not have access to the underlying data generating distribution $p_\dataset(h, \xi)$,
we do have access to a training set $\dataset = \{(h_i, \xi_i)\} \sim p_\dataset(h, \xi)$ that contains $N$ samples from this distribution.
Therefore, we train a neural network velocity field $v_\theta(a, t \mid h)$ using a finite-sample estimate of \cref{eq:ideal-cfm-loss}: \mbox{$\widehat{\mathcal{L}}_\mathrm{CFM}(\weights, \dataset) = \frac{1}{N} \sum_{i=1}^N \EBigSubscript{0.1}{12}{t \sim {U}[0,1]} \EBigSubscript{0.1}{12}{a \sim p_{\xi_i}\left(a \mid t\right)} \big\|v_\weights(a, t \mid h_i) - v_{\xi_i}(a, t)\big\|_2^2$} as shown in \cref{alg:training}.

\textbf{Inference:} While behavior policies are trained to predict sequences of horizon $\Tpred$,
they are usually run in a receding horizon fashion with a potentially different action chunk horizon $\Tchunk \leq \Tpred$~\cite{chi2023diffusion}.
The integration timestep $\Delta t$ is another hyperparameter that controls the granularity of the action sequence.
Therefore, to generate an action chunk, we integrate the velocity field in $t \in [0, \Tchunk / \Tpred]$, producing $\Tchunk / (\Tpred \Delta t)$ many actions.
The action chunk is computed and executed open-loop \ie the neural network $v_\theta$ inputs the same observation history $h_\mathrm{chunk}$ for all integration steps.
Importantly, we are able to stream and execute actions on the robot as soon as they are computed (see \cref{alg:inference}, \cref{line:algo-inference-execute}).
In contrast, diffusion$/$flow policies must wait for the inner loop to complete before executing any actions.

\textbf{Deterministic execution at test time:} Our learning framework suggests the initial action be sampled from $a_0 \sim \N\left(a_0 \mid \aprev, \sigma_0^2\right)$ (see \cref{eq:flow-integration,eq:stabilizing-conditional-flow-velocity-field}).
However, during inference time, we avoid adding noise to actions by setting $\sigma_0 = 0$ to produce deterministic behavior.
We do so because the ability to represent multi-modal distributions is primarily motivated by the need to prevent ``averaging" distinct but valid behaviors of the same task~\cite{chi2023diffusion}.
While representing multi-modality is crucial during training,
the learned policy can be run deterministically at test time without loss in performance.
For example, ACT~\cite{zhao2023learning} sets its variance parameter to zero at test time to produce deterministic behavior.
In \cref{app:stochastic-sfp}, we present a variant of streaming flow policy in an extended state space that decouples stochasticity into additional latent variables.
This variant allows us to sample multiple modes of the trajectory distribution at test time without adding noise to actions.
However, we found that simply setting $\sigma_0 = 0$ at test time works better in practice; therefore we follow this strategy in all our experiments.

\textbf{Imitating actions vs. states:} When training trajectories correspond to \textit{actions}, we start integration of the current action chunk from the most recently generated action in the previous chunk.
Streaming flow policy can also be used to imitate robot \textit{state} trajectories when a controller is available that can closely track desired states.
It is especially suited for state imitation because we can start integration of the current state chunk from the \textit{current robot state} that is accurately measured by proprioceptive sensors.
Therefore, streaming flow policy is able to leverage state feedback in two ways: in the history $h$ and the initialization $a_0$ for flow integration.
This reduces error in the generated trajectory.

{
\newcommand{\emphasis}[1]{#1}
\newcommand{\hthickness}{1.3pt}
\newcommand{\vthickness}{1.3pt}
\newcommand{\NA}{\it \texttt{Not Applicable} \cellcolor{gray!20}}
\newcolumntype{I}{!{\vrule width \vthickness}}
\newcolumntype{|}{!{\vrule width 0.8pt}}
\newcolumntype{:}{!{\vrule width 0pt}}

\renewcommand{\arraystretch}{1.122}  

\begin{table*}[t!]

    \setlength\tabcolsep{3pt} 
    \small
    \centerline{
    \begin{tabular}{Ic|cIc|c|cIc|cI}
        \Xhline{\hthickness}
        \multicolumn{2}{IcI}{
            \multirow{3}{*}[-0.5ex]{
                \vspace*{5.5em}
                \includegraphics[width=0.065\textwidth]{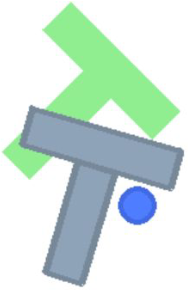}
            }
        } & 
        \multicolumn{3}{cI}{Push-T with state input} & \multicolumn{2}{cI}{Push-T with image input}\\
        \Xcline{3-7}{\hthickness}
        \multicolumn{2}{IcI}{} & \rule{0pt}{3.6ex}{\makecell{State imitation\\Avg/Max scores}} & 
        {\makecell{Action imitation\\Avg/Max scores}} & Latency & 
        {\makecell{State imitation\\Avg/Max scores}} &
        Latency \\
        \cline{3-7}
        \multicolumn{2}{IcI}{} & {$\uparrow$} & 
        {$\uparrow$} & $\downarrow$ & {$\uparrow$} & $\downarrow$ \\
            \Xhline{\hthickness}
            \rowcolor{\redcolor}
            1 & DP~\cite{chi2023diffusion}: 100 DDPM steps & 92.9\% $/$ 94.4\% & 90.7\% $/$ 92.8\% & 40.2 ms & \textbf{87.0\%} $/$ 90.1\% & 127.2 ms \\
            \cline{1-7}
            \rowcolor{\redcolor}
            2 & DP~\cite{chi2023diffusion}: \ \ 10 DDIM steps & 87.0\% $/$ 89.0\% & 81.4\% $/$ 85.3\% & 04.4 ms & 85.3\% $/$ \textbf{91.5\%} & 10.4 ms \\
            \cline{1-7}
            \rowcolor{\redcolor}
            3 & Flow matching policy~\cite{zhang2024affordance} & 80.6\% $/$ 82.6\% & 80.6\% $/$ 82.6\% & 05.8 ms & 71.0\% $/$ 72.0\% & 12.9 ms \\
            \cline{1-7}
            \rowcolor{\redcolor}
            4 & Streaming DP~\cite{hoeg2024streaming} & 87.5\% $/$ 91.4\% & 84.2\% $/$ 87.0\% & 26.7 ms & 84.7\% $/$ 87.1\% & 77.7 ms \\
            \Xhline{\hthickness}
            \rowcolor{\bluecolor}
            5 & SFP \textit{without} {stabilization} & {84.0\% $/$ 86.4\%} & 81.8\% $/$ 93.2\% & {\bf 03.5 ms} & 73.9\% $/$ 77.5\% & \textbf{08.8 ms} \\
            \Xhline{\hthickness}
            \rowcolor{\greencolor}
            6 & \textbf{SFP (Ours)} & {\bf 95.1\% $/$ 96.0\%} & \textbf{91.7\%} $/$ \textbf{93.7\%} & {\bf 03.5 ms} & 83.9\% $/$ 84.8\% & \textbf{08.8 ms} \\
            \Xhline{\hthickness}
        \end{tabular}
    }
    \vspace{-0.4em}
    \caption{
        \centering
        \small
        Imitation learning accuracy on the Push-T~\cite{chi2023diffusion} dataset.\\
        \greenbox{2pt}~Our method (in green) compared against \redbox{2pt}~baselines (in red) $/$ and \bluebox{2pt}~ablations (in blue). See text for details.
    }
    \label{table:results-pusht}
    \vspace{-1.2em}
\end{table*}
}
{
\newcommand{\emphasis}[1]{#1}
\newcommand{\hthickness}{1.3pt}
\newcommand{\vthickness}{1.3pt}
\newcommand{\NA}{\it \texttt{Not Applicable} \cellcolor{gray!20}}
\newcolumntype{I}{!{\vrule width \vthickness}}
\newcolumntype{|}{!{\vrule width 0.8pt}}
\newcolumntype{:}{!{\vrule width 0pt}}

\renewcommand{\arraystretch}{1.122}  

\begin{table*}[h!]

    \setlength\tabcolsep{7pt} 
    \small
    \centerline{
    \begin{tabular}{Ic|cIcIcIcIcI}
        \Xhline{\hthickness}
        \multicolumn{2}{IcI}{
            \multirow{2}{*}[2.5ex]{
                \hspace{-1.1em}
                \includegraphics[width=0.107\textwidth]{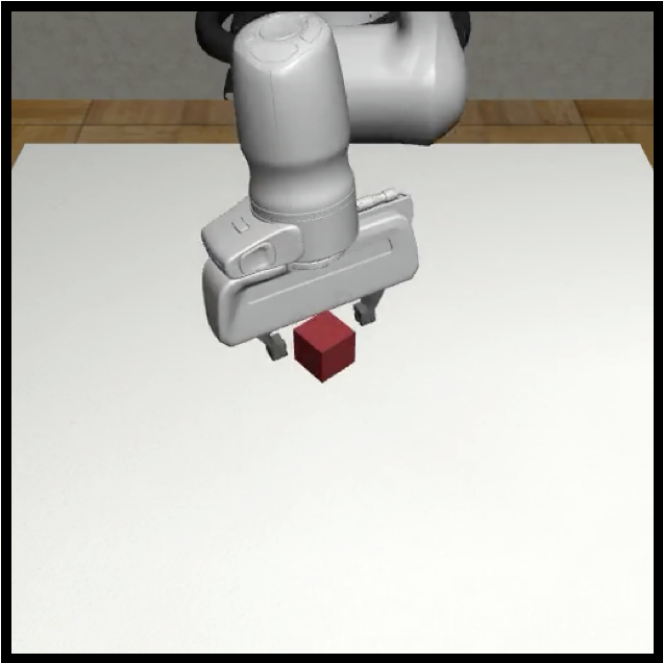}
                \includegraphics[width=0.107\textwidth]{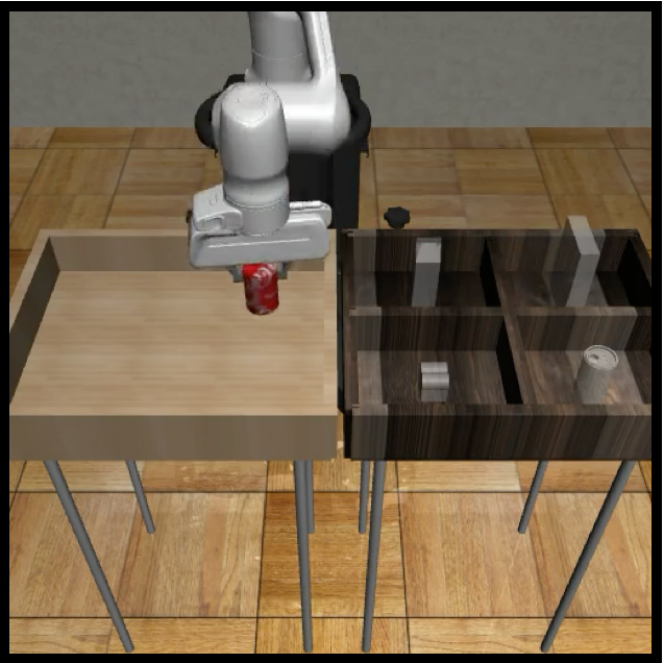}
                \includegraphics[width=0.107\textwidth]{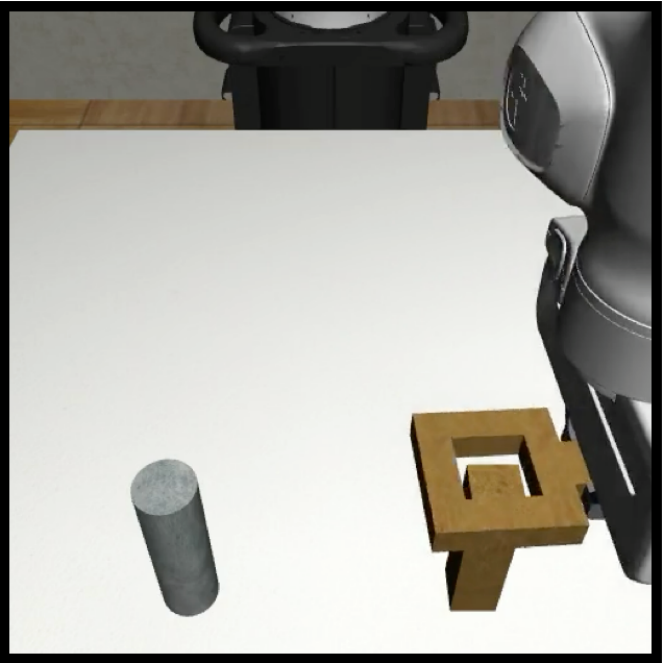}
                \hspace{-1.1em}
            }
        }
        &
        \multicolumn{1}{cI}{\makecell{\textbf{RoboMimic Lift}\\Action imitation\\Avg/Max scores}} & \multicolumn{1}{cI}{\makecell{\textbf{RoboMimic Can}\\Action imitation\\Avg/Max scores}} & \multicolumn{1}{cI}{\makecell{\textbf{RoboMimic Square}\\Action imitation\\Avg/Max scores}} & \multicolumn{1}{cI}{\textbf{Latency}}\\
        \cline{3-6}
        \multicolumn{2}{IcI}{} & 
        {$\uparrow$} & {$\uparrow$} & {$\uparrow$} & {$\downarrow$} \\
            \Xhline{\hthickness}
            \rowcolor{\redcolor}
            1 & DP~\cite{chi2023diffusion}: 100 DDPM steps & \textbf{100.0\%} $/$ \textbf{100.0\%} & 94.0\% $/$ 98.0\% & 77.2\% $/$ \textbf{84.0\%} & 53.4 ms \\
            \cline{1-6}
            \rowcolor{\redcolor}
            2 & DP~\cite{chi2023diffusion}: \ \ 10 DDIM steps & \textbf{100.0\%} $/$ \textbf{100.0\%} & 94.8\% $/$ 98.0\% & 76.0\% $/$ 82.0\% & 5.8 ms \\
            \cline{1-6}
            \rowcolor{\redcolor}
            3 & Flow matching policy~\cite{zhang2024affordance} & ~~99.2\% $/$ \textbf{100.0\%} & 66.0\% $/$ 80.0\% & 54.0\% $/$ 56.0\% & 4.8 ms \\
            \cline{1-6}
            \rowcolor{\redcolor}
            4 & Streaming DP~\cite{hoeg2024streaming} & ~~98.8\% $/$ \textbf{100.0\%} & 96.8\% $/$ 98.0\% & 77.6\% $/$ 82.0\% & 30.3 ms \\
            \Xhline{\hthickness}
            \rowcolor{\bluecolor}
            5 & SFP \textit{without} {stabilization} & ~~99.6\% $/$ \textbf{100.0\%} & 90.0\% $/$ 92.0\% & 53.2\% $/$ 60.0\% & \textbf{4.5 ms} \\
            \Xhline{\hthickness}
            \rowcolor{\greencolor}
            6 & \textbf{SFP (Ours)} & \textbf{100.0\%} $/$ \textbf{100.0\%} & \textbf{98.4\%} $/$ \textbf{100.0\%} & \textbf{78.0}\% $/$ \textbf{84.0}\% & \textbf{4.5 ms} \\
            \Xhline{\hthickness}
        \end{tabular}
    }
    \vspace{-0.4em}
    \caption{
        \centering
        \small
        Imitation learning accuracy on RoboMimic~\cite{mandlekar2021what} environment.\\
        \greenbox{2pt}~Our method (in green) compared against \redbox{2pt}~baselines (in red) $/$ and \bluebox{2pt}~ablations (in blue). See text for details.
    }
    \label{table:results-robomimic}
    \vspace{-1.5em}
\end{table*}
}
\section{Experiments}
\label{sec:experiments}

\noindent We evaluate streaming flow policy on two imitation learning benchmarks: the Push-T environment~\cite{florence2022implicit,chi2023diffusion}, and RoboMimic~\cite{mandlekar2021what}\add{, and perform real-world experiments on a Franka Research 3 robot arm.}

We compare our method (in green \greenbox{1pt}) against 4 baselines (in red \redbox{1pt}): Row 1 (DP): standard diffusion policy~\cite{chi2023diffusion} that uses 100 DDPM~\cite{ho2020denoising} steps, Row 2 (DP): a faster version of diffusion policy that uses 10 DDIM~\cite{songdenoising} steps, Row 3: conventional flow matching-based policy~\cite{zhang2024affordance} and Row 4: Streaming Diffusion Policy~\cite{hoeg2024streaming}, a recent method that runs diffusion policy in a streaming manner (see \cref{sec:related-work}).
We also compare against streaming flow policy that does not construct stabilizing flows during training, \ie uses $k=0$ (in blue \bluebox{1pt}).
This ablation is designed to measure the contribution of stabilization to task performance.
Following \citet{chi2023diffusion}, we report the average score for the 5 best checkpoints, the best score across all checkpoints, and the average latency per action for each method.

In \cref{table:results-pusht}, we report results on the Push-T environment: a simulated 2D planar pushing task where the robot state is the 2-D position of the cylindrical pusher in global coordinates, and actions are 2-D setpoints tracked by a PD controller.
Push-T contains 200 training demonstrations and 50 evaluation episodes.
We perform experiments in two settings: when simulator state is used as observations, and when images are used as observations.
``Action imitation'' is the standard practice of imitating action sequences provided in the benchmark training set.
We also perform experiments with ``state imitation'' (see \cref{sec:learning-marginal-velocity-fields}), where we directly imitate the measured 2D positions of the robot.
Here, we use the known ground-truth robot position at the beginning of each action chunk to as a starting point for velocity field integration.
In \cref{table:results-robomimic}, we report results on the RoboMimic environment, specifically the ``lift'' and ``can'' tasks, with state inputs.
Both tasks involve predicting sequences of 6-DOF end-effector poses with respect to a global frame that are tracked by a PD controller.
Each task contains 300 training demonstrations, and 50 evaluation episodes.
The tasks involve picking objects and placing them at specific locations, including picking a square nut and placing it on a rod.

The neural network for streaming flow policy $v_\theta: \A \times [0, 1] \times \H \to \TA$ is structurally similar to diffusion$/$flow policies (\eg ${\epsilon}_\weights: \A^T \times [0, 1] \times \H \to \A^T$) with the only change being the input and output spaces (action space $\A$ vs. action-trajectory space $\A^T$).
Therefore, we are able to re-use existing diffusion$/$flow policy architectures~\cite{chi2023diffusion} by changing the input and output dimension of the network and replacing 1-D temporal convolution/attention layers over action sequences~\cite{janner22aplanning,chi2023diffusion} with a fully connected layer.
Furthermore, due to the reduced dimensionality of the flow sampling space, we found that streaming flow policy is faster to train and has a smaller GPU memory footprint compared to diffusion$/$flow policies.

{
\begin{figure}[t!]
\centerline
{
  \includegraphics[trim=0 0 0 0,clip,width=0.55\textwidth]{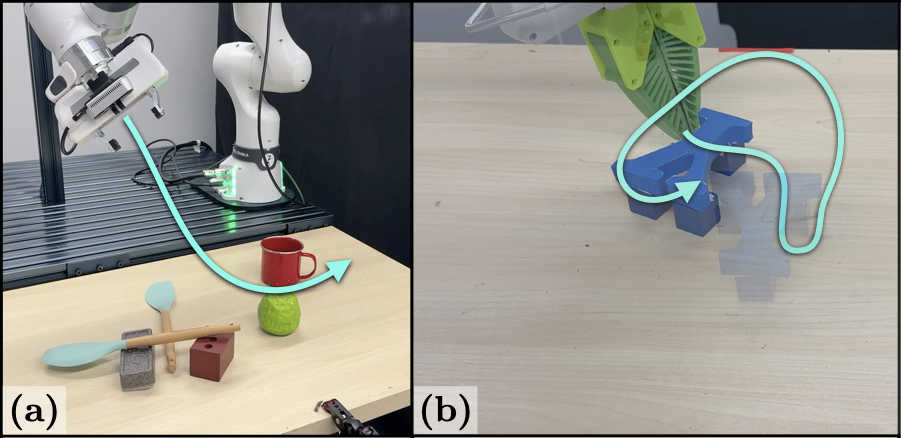}
}
\vspace*{-0.5em}
\captionof{figure}
{
  \small
  Real-world experiments on a 7-DOF Franka Research 3 robot arm on the tasks of
  \textbf{(a)}
  reaching and picking an apple, and
  \textbf{(b)}
  reorienting a block.
  We compare our method against diffusion policy, and find that our method is noticeably faster and produces smoother motion;
  see comparison videos on our \href{\website}{\color{deeppink}project website}.
}
\label{fig:real-world}
\vspace*{-1em}
\end{figure}
}

\add{
We conduct real-world experiments on a Franka Research 3 robot arm with a RealSense D435f depth camera on two tasks: (a) reaching and picking an object, and (b) reorienting a block to a vertical goal orientation (see \cref{fig:real-world}).
We find that streaming flow policy is noticeably faster and produces smoother motion than diffusion policy, as shown in videos on the \href{\website}{\color{deeppink}project website}.
}

\textbf{Conclusions}: Stabilizing flow policy performs comparably to diffusion policy and other baselines in terms of performance on most tasks, while being significantly faster per action.
Furthermore, the reported latency does not even take into account the fact that streaming flow policy can asynchronously run action generation and robot execution in parallel.
In practice, this can avoid delays and jerky robot movements.
Diffusion policy can be sped up by running fewer diffusion steps via DDIM~\cite{songdenoising}.
And flow-matching policy is also faster than diffusion policy. However, their speed seems to come at the cost of sometimes significant reduction in accuracy.
In \cref{app:action-horizon}, we analyze the performance of streaming flow policy as a function of the action chunk horizon $\Tchunk$.

\section{Related work}
\label{sec:related-work}

\noindent\textbf{Learning dynamical systems:}
Our work is closely related to prior work on learning dynamical systems~\cite{khansari2010imitation,khansari2011learning,sochopoulos2024learning,ijspeert2013dynamical,schaal2003computational}. A key difference is that most prior works assume a single, deterministic future trajectory given an input state. However, we focus on learning multi-modal \textit{distributions} over trajectories, which is known to be essential for behavior cloning in robotics~\cite{chi2023diffusion}.
For example, \citet{sochopoulos2024learning} learn a neural ODE~\cite{chen2018neural} by minimizing the distance between the predicted and demonstrated trajectories. This is susceptible to undesirable ``averaging" of distinct behaviors that achieve the same task.
\citet{singh2022multiscale} learn a neural CDE~\cite{kidger2020neural}, motivated by asynchronous multiscale sensor fusion.
Our approach learns a neural ODE endowed with an initial noise distribution that induces a distribution over future trajectories.
This is also known as a continuous normalizing flow~\cite{chen2018neural,grathwohl2018scalable}.
We use the recently proposed flow matching framework~\cite{lipman2023flow} to fit the predicted trajectory distribution to training data.
An important consequence is that our method aggregates ``noised" demonstration trajectories as additional training samples, whereas prior works train only on states directly present in the demonstrations. 
Furthermore, most prior works assume a time-invariant velocity field~\cite{khansari2010imitation,khansari2011learning,sochopoulos2024learning}.
Our velocity field depends on time and allows learning non-Markovian behaviors for tasks like spreading sauce on pizza dough, where the end-effector must rotate a fixed number of times. 
Finally, most works on learning dynamical systems focus on \textit{point-stability} around goal states~\cite{khansari2010imitation,khansari2011learning,sochopoulos2024learning};
we don't assume goal states and construct flows that stabilize around demonstration trajectories.

\noindent\textbf{Flow matching:}
Flow matching~\cite{lipman2023flow} is a recent technique for learning complex, multi-modal distributions that has been used to model images~\cite{lipman2023flow,esser2024scaling,ma2024sit}, videos~\cite{jin2024pyramidal,polyak2025moviegencastmedia}, molecular structures~\cite{jing2024alphafold,bose2023se,klein2023equivariant}, and robot action sequences~\cite{black2410pi0,ye2024efficient,funk2024actionflow,zhang2024affordance,braun2024riemannian}.
However, the flow sampling process starts from Gaussian noise, and the distribution of interest is only modeled at the final timestep $t=1$.
Our insight is to treat the entire flow trajectory as a sample from the target distribution over action sequences.

\noindent\textbf{Streaming Diffusion Policy:}
The work most closely related to ours is Streaming Diffusion Policy~\cite{hoeg2024streaming}, which is an adaptation of of discrete time diffusion policies~\cite{chi2023diffusion}.
Instead of maintaining all actions in the sequence at the same noise level, Streaming Diffusion Policy maintains a rolling buffer of actions with increasing noise levels.
Every diffusion step reduces the noise level of all actions by one, fully de-noising the oldest action in the buffer that can be streamed to the robot.
However, this method requires maintaining an action buffer of the length of the prediction horizon, even if the action horizon is much shorter.
Furthermore, there is an up-front cost to initialize the buffer with increasing noise levels.
The rolling buffer approach has been applied to video prediction~\cite{ruhe2024rolling} and character motion generation~\cite{zhang2024tedi}.
Our method is more economical since it computes only as many actions are streamed to the robot, without requiring a buffer.
We evaluate our method against Streaming Diffusion Policy in \cref{sec:experiments}.

\section{Conclusion}

In this work, we have presented a novel approach to imitation learning that addresses the computational limitations of existing diffusion and flow-matching policies.
Our key contribution is a simplified approach that treats \textit{action trajectories as flow trajectories}. This enables incremental integration of a learned velocity field that allows actions to be streamed to the robot during the flow sampling process. The streaming capability makes our method particularly well-suited for receding horizon policy execution.
Despite the streaming nature of our approach, the flow matching framework guarantees the ability to model multi-modal action trajectories.
By constructing flows that stabilize around demonstration trajectories, we reduce distribution shift and improve imitation learning performance.
Our experimental results demonstrate that streaming flow policy performs comparably to prior imitation learning approaches on benchmark tasks, but enables faster policy execution and tighter sensorimotor loops, making it more practical for reactive, real-world robot control.


\clearpage
\section{Limitations}
\label{sec:limitations}
In this section, we discuss some limitations of our approach.

\subsection{SFP does not match joint distribution, only per-timestep marginal distributions}

Our flow matching framework ensures that the learned distribution over trajectories conditioned on the history matches the training distribution in terms of marginal distributions of actions at each timestep $t \in [0, 1]$.
We however, \textbf{do not guarantee that the \textit{joint} distribution of actions across a trajectory matches the training distribution}.
This is in contrast to diffusion policy, that is able to match the joint distribution since the diffusion model operates in trajectory space $\A^T$.

\begin{figure*}[h!]
  \centerline
  {
    \includegraphics[trim=0 0 0 0,clip,width=1.1\textwidth]{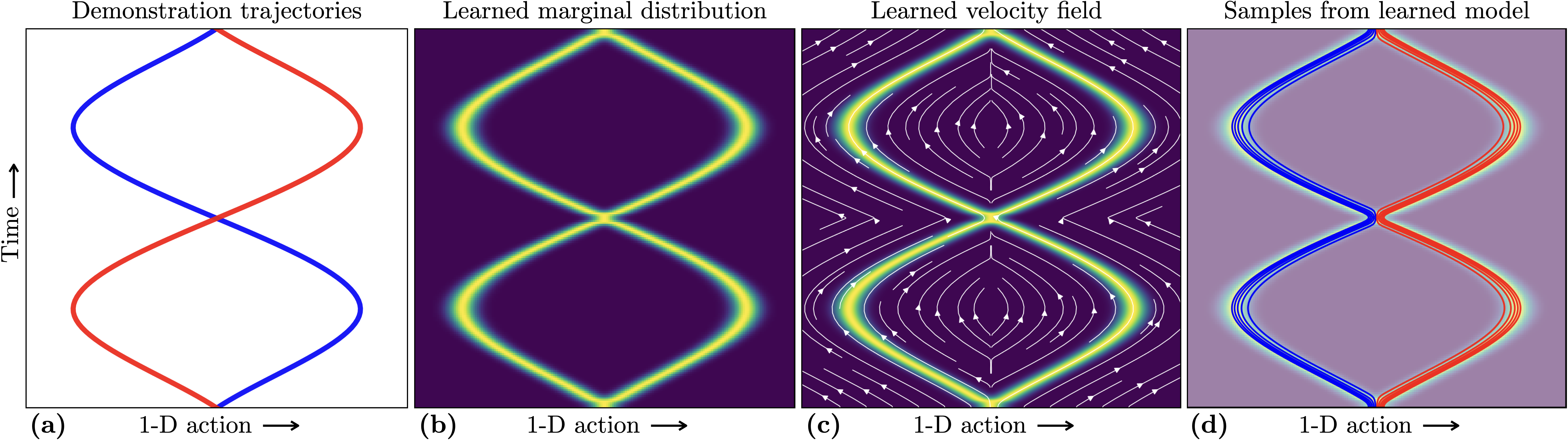}
  }
  \captionof{figure}
  {
    \small
    A toy example illustrating how streaming flow policy matches marginal distribution of actions in the trajectory at all time steps, but not necessarily their joint distribution.
    The $x$-axis represents a 1-D action space, and the $y$-axis represents both trajectory time and flow time. 
    \textbf{(a)}
    The bi-modal training set contains two intersecting demonstration trajectories,
    illustrated in blue and red, with shapes ``S'' and ``\reflectbox{S}'' respectively.
    \textbf{(b)} The marginal distribution of actions at each time step learned by our streaming flow policy.
    The marginal distributions perfectly match the training data.
    \textbf{(c)} The learned velocity flow field $v_\weights(a, t \mid h)$ that yeilds the marginal distributions in (b).
    \textbf{(d)} Trajectories sampled from the learned velocity field. Trajectories that start from $a < 0$ are shown in blue, and those starting from $a > 0$ are shown in red. The sampled trajectories have shapes ``\reflectbox{$3$}'' and ``3'', with equal probability. These shapes are different from the shapes ``S'' and ``\reflectbox{S}'' in the training distribution, although their margin distributions are identical.
  }
  \label{fig:limitations-det}
\end{figure*}

\begin{figure}[t]
  \centerline
  {
    \includegraphics[trim=0 0 0 0,clip,width=0.6\columnwidth]{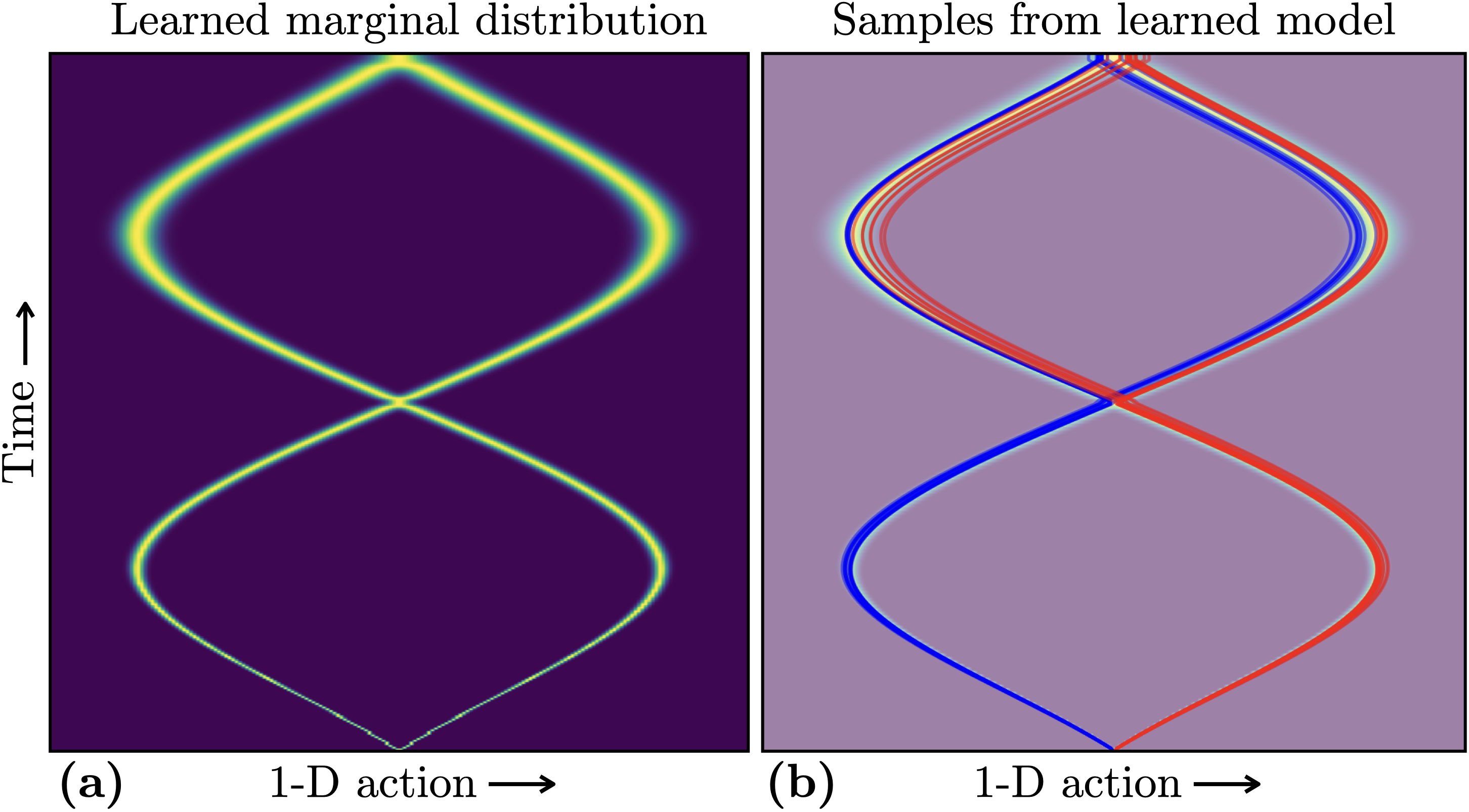}
  }
  \captionof{figure}
  {
    \small
    Different variants of streaming flow policy can produce different joint distributions of actions that are consistent with the marginal distributions in the training data.
    This example is produced using the latent-variable   version of streaming flow policy, described in \cref{app:stochastic-sfp}.
    \textbf{(a)}
    The marginal distribution of actions at each time step learned by the streaming flow policy matches the training data.
    \textbf{(b)} Samples from the trained policy produces \textit{four} modes with shapes ``S'', ``\reflectbox{S}'', ``\reflectbox{3}'' and ``3'', whereas the training data contains only two modes with shapes ``S'' and ``\reflectbox{S}''.
  }
  \label{fig:limitations-stoc}
\end{figure}

\cref{fig:limitations-det,fig:limitations-stoc} illustrate a toy example where streaming flow policy matches marginal distributions but not the joint distribution.
The $x$-axis represents 1-D robot actions, and the $y$-axis represents flow time ($t \in [0, 1]$).
\cref{fig:limitations-det}a shows two trajectories in blue and red, of shapes ``S'' and ``\reflectbox{S}'' respectively.
The trajectories intersect at $t = 0.5$.
The learned flow field is shown in \cref{fig:limitations-det}c, and the induced marginal distribution over actions is shown in \cref{fig:limitations-det}d.
The marginal distribution of actions matches the training distribution at each $t \in [0, 1]$.
Trajectories sampled from the flow field are shown in \cref{fig:limitations-det}d.
The trajectory distribution contains two modes of equal probability: trajectories that always lie either in $a < 0$ (shown in blue), or in $a > 0$ (shown in red).
The shapes formed by sampled trajectories --- ``\reflectbox{3}" and ``3" respectively --- do not match the shapes of trajectories in the training data.

A similar phenomenon is illustrated in \cref{fig:limitations-stoc} using the latent-space variant of streaming flow policy (see \cref{app:stochastic-sfp}) trained on the same dataset of intersecting trajectories.
While the marginal distribution of actions again matches with the training distribution, the trajectories contain \textit{four} modes, with shapes ``S'', ``\reflectbox{S}'', ``\reflectbox{3}'' and ``3''.
Note that the per-timestep marginal distributions over actions still match the training data.


\subsection{Streaming flow policies exhibit compositionality}

\textbf{The loss of fidelity to the joint distribution} is a potential weakness of our framework.
Therefore, this framework may not be the right choice when learning the correct joint distributions is crucial.
However, another perspective is to think of our method as providing \textit{compositionality} over training demonstrations.
The sampled trajectories can be composed of pieces across the training data.

For many robotics tasks, compositionality might be both valid and desirable.
For example, in quasi-static tasks where the robot moves slowly, if two demonstration trajectories are valid, then the compositions across these trajectories are often also valid.
Under this assumption, compositionality allows the flow model to learn many valid combinations of partial trajectories with fewer demonstrations.

What constraints on trajectories reflected in the training data can streaming flow policy learn?
Streaming flow policy is \textbf{unable to capture global constraints} that can only be represented in the joint distribution.
However, it can learn certain local constraints.

\subsection{SFPs can learn arbitrary position constraints}

Robot actions $a \in Q \subseteq \A$ may be constrained to lie in a subset $Q \subseteq \A$.
For example, $Q$ may reflect joint limits of a robot arm.
Then, a well-trained streaming flow policy should learn this constraint as well.

To see why, consider \cref{eq:averaged-marginal-distributions} which states that the learned marginal density of actions $p^*(a \mid t, h) = \int_\xi p_\xi(a \mid t)\,p_\dataset(\xi \mid h) \, \dd\xi$ at time $t$ is a weighted average of marginal densities of conditional flows $p_\xi(a \mid t)$.
Recall that we construct $p_\xi(a \mid t)$ to be narrow Gaussian tubes around demonstration trajectories $\xi$.
Assume that the thickness of the Gaussian tube is sufficiently small that $a \notin Q \implies p_\xi(a \mid t) < \epsilon$, for some small $\epsilon > 0$ and for all $\xi, t$.
Then we have from \cref{eq:averaged-marginal-distributions} that $p_\xi(a \mid t) < \epsilon \implies p^*(a \mid t, h) < \epsilon$ for all $t \in [0, 1]$.
Therefore, the probability of sampling an action $a$ that violates the constraint $Q$ is extremely low.

\subsection{SFPs can learn convex velocity constraints}

Theorem 2 of \citet{lipman2023flow} implies that the minimizer of the conditional flow matching loss $v^* \coloneq \arg \min_v \lcfm(v, p_\dataset)$ has the following form:
{
\newcommand{\denom}{\int_{\xi'} p_\dataset(\xi' \mid h)\,p_{\xi'}(a \mid t) \, \dd\xi'}
\begin{align}
    v^*(a, t \mid h)
        &= \int_\xi v_\xi(a, t) \, \underbrace{\frac{p_\dataset(\xi \mid h)\,p_\xi(a \mid t)}{\denom}}_{\displaystyle p_\dataset(\xi \mid a, t, h)} \, \dd\xi \nonumber\\ 
        &= \int_\xi v_\xi(a, t) \, p_\dataset(\xi \mid a, t, h) \, \dd\xi
        \label{eq:velocity-convex-combination}\\
        &\approx \int_\xi \dot{\xi}(t) \, p_\dataset(\xi \mid a, t, h) \, \dd\xi \ \text(\text{\small assuming } k \approx 0)\nonumber
    \end{align}
}
Intuitively, the target velocity field $v^*$ at $(a, t)$ is a weighted average of conditional flow velocities $v_\xi(a, t)$ over demonstrations $\xi$.
The weight for $\xi$ is the Bayesian posterior probability of $\xi$, where the prior probability $p_\dataset(\xi \mid h)$ is the probability of $\xi$ given $h$ in the training distribution, and the likelihood $p_\xi(a \mid t)$ is the probability that the conditional flow around $\xi$ generates $a$ at time $t$.

Under sufficiently small values of $k$, we have from \cref{eq:stabilizing-conditional-flow-velocity-field} that $v_\xi(a, t) \approx \dot{\xi}(t)$.
Note that $v^*$ is then a convex combination of demonstration velocities $\dot{\xi}(t)$.
Consider convex constraints over velocities $\dot{\xi}(t) \in C$ \ie $\dot{\xi}(t)$ is constrained to lie in a convex set $C$ for all $\xi$ with non-zero support $p_\dataset(\xi) > 0$ and for all $t \in [0, 1]$.
This is the case, for example, when robot joint velocities lie in a closed interval $[v_\mathrm{min}, v_\mathrm{max}]$.
Then, \cref{eq:velocity-convex-combination} implies that $v^*$ also lies in $C$.



\bibliography{main.bib}


\clearpage
\begin{appendices}
{
    \nolinenumbers{
    \centering
    {\fontsize{22pt}{22pt}\selectfont Appendix}\\
    \vspace{1.5em}
    {
      \LARGE \bf {
        Streaming Flow Policy
      }\\\vspace{0.1em}
      {
        \fontsize{15}{15}\normalfont
        \begin{center}
          Simplifying diffusion$/$flow-matching policies by\\
          treating action trajectories as flow trajectories
        \end{center}
      }
    }
    \vspace{0.1em}
  }
}

\section{Proof of Theorem 1}
\label{app:proof-theorem-1}

Integrating learned velocity fields can suffer from drift since errors accumulate during integration. We adding a stabilization term, we can correct deviations from the demonstration trajectory.
The stabilizing velocity field is:
\begin{equation}
    v_\xi(a, t) = \underbrace{-k(a - \xi(t))}_{\text{Stabilization term}} + \underbrace{\dot{\xi}(t)}_{\text{Path velocity}}
\end{equation}
where $k > 0$ is the stabilizing gain. This results in exponential convergence to the demonstration:
\begin{samepage}
\begin{align}
    &\frac{d}{dt}(a - \xi(t)) = -k(a - \xi(t)) \\
    \implies &\frac{1}{a - \xi(t)} \frac{d}{dt}(a - \xi(t)) = -k \\
    \implies &\frac{d}{dt} \log(a - \xi(t)) = -k \\
    \implies &\log(a - \xi(t))\Big\vert_0^t = -\int_0^t k dt \\
    \implies &\log\frac{a(t) - \xi(t)}{a_0 - \xi(0)} = -kt \\
    \implies &a(t) = \xi(t) + (a_0 - \xi(0))e^{-kt}
\end{align}
\end{samepage}
Since $a_0 \sim \N\left(\xi(0), \sigma_0^2\right)$ (see~\cref{eq:flow-integration}), and $a(t)$ is linear in $a_0$, we have by linearity of Gaussian distributions that:
\begin{equation}
    p_\xi(a \mid t) = \mathcal{N}(a \bigmid \xi(t), \sigma_0^2 e^{-2kt})
\end{equation}

$\square$

\begin{figure*}[ht]
  \centerline
  {
    \includegraphics[trim=0 0 0 0,clip,width=0.9\textwidth]{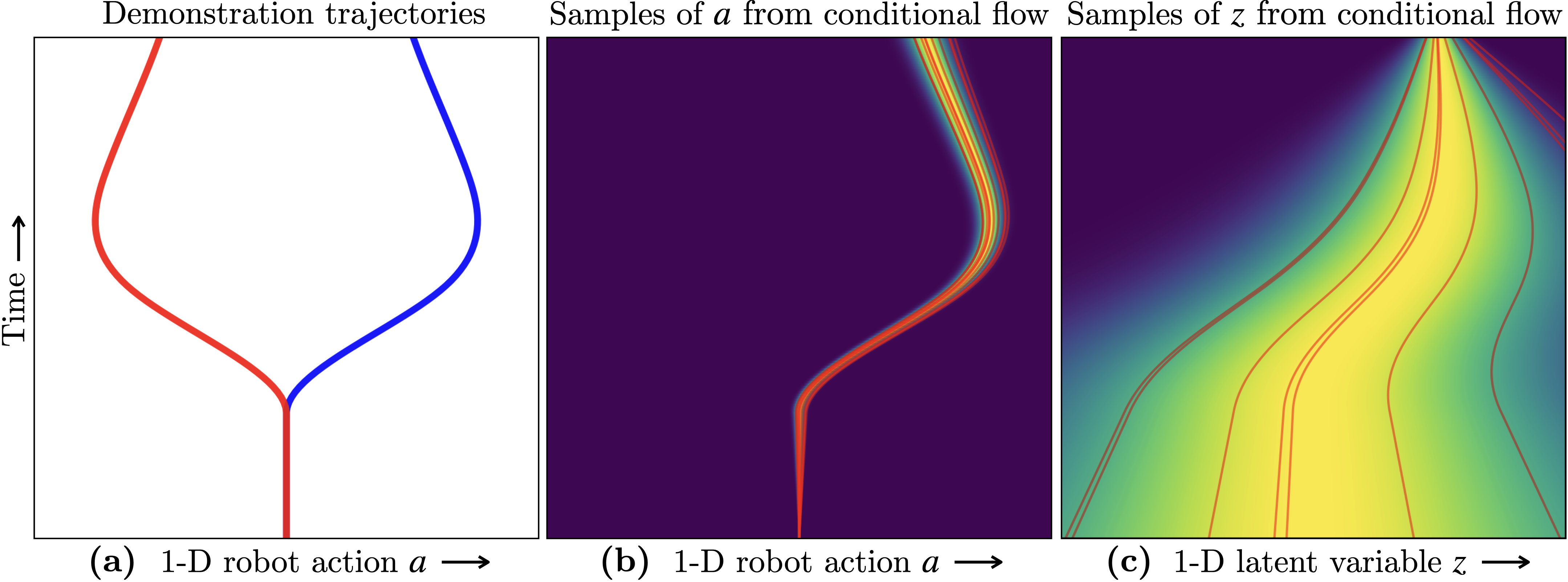}
  }
  \captionof{figure}
  {
    Constructing a conditional flow using auxiliary stochastic latent variables instead of adding noise to actions.
    In this toy example, the $x$-axis represents a 1-D action space, and the $y$-axis represents both trajectory time and flow time.
    \textbf{(a)}
    A toy bi-modal training set contains two trajectories shown in red and blue; the same as in \cref{fig:pull-figure}a.
    Given a demonstration trajectory $\xi$ from the training set (\eg the demonstration in blue), we design a velocity field $v_\xi(a, z, t)$ that takes as input time $t \in [0, 1]$, the action $a$ at time $t$, as well as an additional latent variable $z$.
    The latent variable is responsible for injecting noise into the flow sampling process, allowing the initial action $a(0)$ to be deterministically set to the initial action $\xi(0)$ of the demonstration.
    The latent variable $z(0) \sim \N(0, 1)$ is sampled from the standard normal distribution at the beginning of the flow process, similar to conventional diffusion$/$flow policies.
    The velocity field $v_\xi(a, z, t)$ generates trajectories in an extended sample space
    $[0, 1] \to \A^2$
    where $a$ and $z$ are correlated and co-evolve with time.
    \textbf{(b, c)} Shows the marginal distribution of actions $a(t)$ and the latent variable $z(t)$, respectively, at each time step.
    Overlaid in red are the $a$- and $z$- projections, respectively, of trajectories sampled from the velocity field.
    The action evolves in a narrow Gaussian tube around the demonstration, while the latent variable starts from $\N(0, 1)$ at $t=0$ and converges to the demonstration trajectory at $t=1$; see \cref{app:stochastic-sfp} for a full description of the velocity field.
  }
  \label{fig:stochastic-sfp-conditional}
\end{figure*}

\section{Decoupling stochasticity via latent variables}
\label{app:stochastic-sfp}

In order to learn multi-modal distributions during training, streaming flow policy as introduced in \cref{sec:conditional-velocity-field} requires a small amount of Gaussian noise added to the initial action.
However, we wish to avoid adding noise to actions at test time.
We now present a variant of streaming flow policy in an extended state space by introducing a latent variable $z \in \A$.
The latent variable $z$ decouples stochasticity from the flow trajectory, allowing us to sample multiple modes of the trajectory distribution at test time
while deterministically starting the sampling process from the most recently generated action.
{
\newcommand{\emphasis}[1]{#1}
\newcommand{\hthickness}{1.1pt}
\newcommand{\vthickness}{1.1pt}
\newcolumntype{I}{!{\vrule width \vthickness}}
\newcolumntype{|}{!{\vrule width 0.8pt}}
\newcolumntype{:}{!{\vrule width 0pt}}
\renewcommand{\arraystretch}{1.2}  
\newcolumntype{M}[1]{>{\centering\arraybackslash}m{#1}}
\begin{table}[h!]
    \setlength\tabcolsep{6pt}
    \centerline
    {
    \begin{tabular}{IM{1.4cm}|M{6.5cm}|M{1.25cm}I}
     \Xhline{\hthickness}
     $\sigma_0$ & Initial standard deviation & $\R^+$ \\
     \hline
     $\sigma_1$ & Final standard deviation & $\R^+$ \\
     \hline
     $k$ & Stabilizing gain & $\Rp$ \\
     \Xhline{\hthickness}
     $\sigma_r$ & \vphantom{$\Big\vert$} Residual standard deviation = $\sqrt{\sigma_1^2 - \sigma_0^2 e^{-2k}}$ & $\Rp$ \\
     \Xhline{\hthickness}
    \end{tabular}
    }
    \vspace{0.4em}
    \caption{
        Hyperparameters used in the stochastic variant of streaming flow policy that uses stochastic latent variables.
    }
    \label{table:stochastic-sfp-notation}
\end{table}
}
We now define a conditional flow in the extended state space $(a, z) \in \A^2$.
We define the initial distribution by sampling $a_0$ and $z_0$ independently.
$a_0$ is sampled from a vanishingly narrow Gaussian distribution centered at the initial action of the demonstration trajectory $\xi(0)$, but with a extremely small variance $\sigma_0 \approx 0$.
$z_0$ is sampled from a standard normal distribution, similar to standard diffusion models~\cite{ho2020denoising} and flow matching~\cite{lipman2023flow}.
\begin{samepage}
\begin{align}
    \intertext{\centering\underline{Initial sample}}
    z_0 &\sim \N(0, I) \\
    a_0 &\sim \N(\xi(0), \sigma_0^2)
\end{align}
\end{samepage}
We assume hyperparameters $\sigma_0$, $\sigma_1$ and $k$.
They correspond to the initial and final standard deviations of the action variable $a$ in the conditional flow.
$k$ is the stabilizing gain.
Furthermore, we constrain them such that $\sigma_1 \geq \sigma_0 e^{-k}$.
Then, let us define $\sigma_r \coloneq \sqrt{\sigma_1^2 - \sigma_0^2e^{-2k}}$.
Then we construct the joint \textit{flow trajectories} of $(a, z)$ starting from $(a(0), z(0))$ as:
\begin{align}
    \intertext{\centering\underline{Flow trajectory diffeomorphism}}
    \begin{aligned}
    a(t \mid \xi, a_0, z_0) &= \xi(t) + \left(a_0 - \xi(0)\right) e^{-kt} + (\sigma_r t)z_0\\
    z(t \mid \xi, a_0, z_0) &= (1 - (1 - \sigma_1)t)z_0 + t \xi(t)
    \end{aligned}
    \label{eqn:ssfp-flow-diffeomorphism}
\end{align}
The flow is a diffeomorphism from $\A^2$ to $\A^2$ for every $t \in [0, 1]$.

Note that $a(0 \mid \xi, a_0, z_0) = a_0$ and $z(0 \mid \xi, a_0, z_0) = z_0$, so the diffeomorphism is identity at $t=0$.
The marginal distribution at $t=1$ for $a$ and $z$ is given by $a(1 \mid \xi) \sim \N(\xi(1), \sigma_1^2)$ and $z(1 \mid \xi) \sim \N(\xi(1), \sigma_1^2)$.

Intuitively, the variable $a$ follows the shape of the action trajectory $\xi(t)$ with an error starting from $a_0 - \xi(0)$ and decreasing with an exponential factor due to the stabilizing gain.
However, it uses the sampled noise variable $z_0 \sim \N(0, I)$ to increase the standard deviation from $\sigma_0$ around $\xi(0)$ to $\sigma_1$ around $\xi(1)$.
This is done in order to sample different modes of the trajectory distribution at test time.
On the other hand, the latent variable $z$ starts from the random sample $z_0 \sim \N(0, I)$ but continuously moves closer to the demonstration trajectory $\xi(t)$, reducing its variance from $1$ to $\sigma_1$.

Since $(a, z)$ at time $t$ is a linear transformation of $(q_0, z_0)$, the joint distribution of $(a, z)$ at every timestep is a Gaussian given by:
\begin{align}
    \intertext{\centering\underline{Joint distribution of $(a, z)$ at each timestep
    }}
    \begin{bmatrix}a\\z\end{bmatrix} =& \underbrace{\begin{bmatrix} e^{-kt} & \sigma_r t \\ 0 & 1-(1-\sigma_1)t \end{bmatrix}}_{\displaystyle A}\begin{bmatrix}a_0\\z_0\end{bmatrix} + \underbrace{\begin{bmatrix}\xi(t) - \xi(0)e^{-kt}\\ t\xi{t}\end{bmatrix}}_{\displaystyle b}\\
    p_\xi(a, z \mid t) =& ~\N \left( A \mu_0 + b \,,\, A \Sigma_0 A^T\right)\\
    =& ~\N \left( \begin{bmatrix}\phantom{t}\xi(t)\\ t\xi(t) \end{bmatrix}, \begin{bmatrix} \Sigma_{11} & \Sigma_{12} \\ \Sigma_{12} & \Sigma_{22}\end{bmatrix}\right) \text{ where} \label{eqn:ssfp-joint-distribution}\\
    &~~\Sigma_{11} = \sigma_0^2 e^{-2kt} + \sigma_r^2 t^2 \\
    &~~\Sigma_{12} = \sigma_r t \left(1 - (1-\sigma_1)t\right)\\
    &~~\Sigma_{22} = \left(1 - (1-\sigma_1)t\right)^2
\end{align}
Note that $\mu_0 = \begin{bmatrix}\xi(0)\\0 \end{bmatrix}$ and $\Sigma_0 = \begin{bmatrix} \sigma_0^2 & 0 \\ 0 & 1\end{bmatrix}$.

\begin{figure*}[h]
  \centerline
  {
    \includegraphics[trim=0 0 0 0,clip,width=1.15\textwidth]{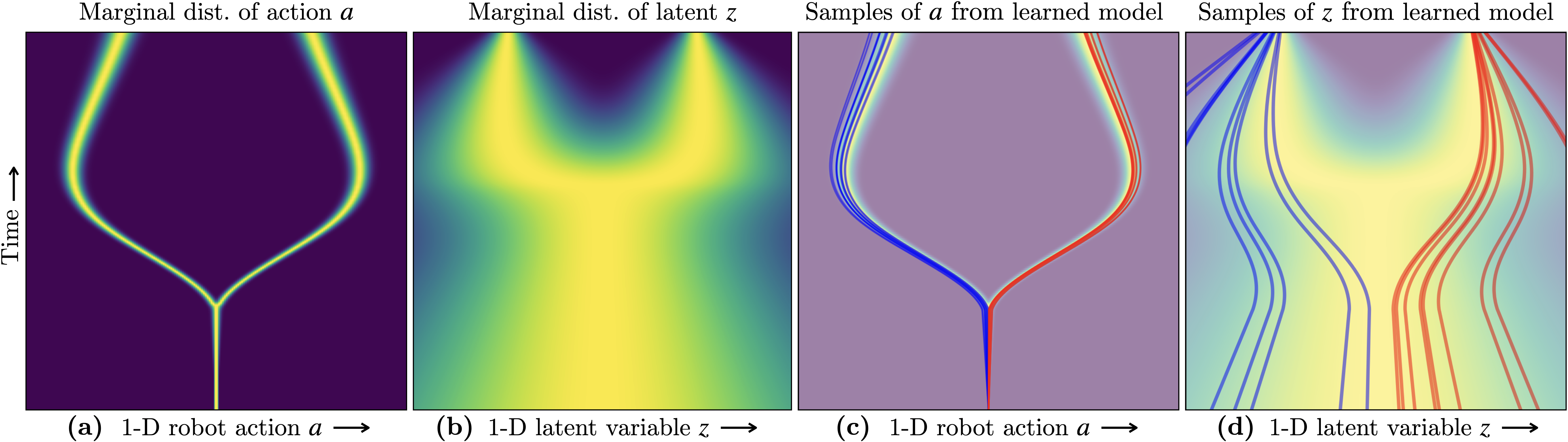}
  }
  \captionof{figure}
  {
    The marginal velocity flow field $v_\weights(a, z, t \mid h)$ learned using the flow construction in \cref{fig:stochastic-sfp-conditional}.
    \textbf{(a, b)}
    shows the marginal distribution of actions $a(t)$ and the latent variable $z(t)$, respectively, at each time step under the learned velocity field.
    \textbf{(c, d)}
    Shows the $a$- and $z$- projections, respectively, of trajectories sampled from the learned velocity field.
    By construction, $a(0)$ deterministically starts from the most recently generated action, whereas $z(0)$ is sampled from $\N(0, 1)$.
    Trajectories starting with $z(0) < 0$ are shown in blue, and those with $z(0) > 0$ are shown in red.
    The main takeaway is that in \textbf{(c)}, even though all samples deterministically start from the same initial action (\ie the most recently generated action), they evolve in a stochastic manner that covers both modes of the training distribution.
    This is possible because the stochastic latent variable $z$ is correlated with $a$, and the initial random sample $z(0) \sim \N(0, 1)$ informs the direction $a$ evolves in.
  }
  \label{fig:stochastic-sfp-marginal}
\end{figure*}

Since the flow is a diffeomorphism, we can invert it and express $(a_0, z_0)$ as a function of $(a(t), z(t))$:
\begin{align}
    \intertext{\centering\underline{Inverse of the flow diffeomorphism}}
    \begin{aligned}
    z_0 &= \frac{z - t \xi(t)}{1 - (1 - \sigma_1)t} \\
    a_0 &= \xi(0) + \left(a - \xi(t) - (\sigma_r t) z_0\right)e^{kt}
    \end{aligned}
    \label{eqn:ssfp-flow-inverse}
\end{align}
At time $t$, the velocity of the trajectory starting from $(a_0, z_0)$ can be obtained by differentiating the flow diffeomorphism in \cref{eqn:ssfp-flow-diffeomorphism} with respect to $t$:
\begin{align}
    \intertext{\centering\underline{Velocity in terms of $(a_0, z_0)$}}
    \begin{aligned}
    \dot{a}(t \mid \xi, a_0, z_0) &= \dot{\xi}(t) - k\left(a_0 - \xi(0)\right) e^{-kt} + \sigma_r z_0 \\
    \dot{z}(t \mid \xi, a_0, z_0) &= \xi(t) + t \dot{\xi}(t) - (1 - \sigma_1)z_0
    \end{aligned}
    \label{eqn:ssfp-flow-velocity}
\end{align}
The flow induces a velocity field at every $(a, z, t)$. The conditional velocity field $v_\weights(a, z, t \mid h)$ by first inverting the flow transformation as shown in \cref{eqn:ssfp-flow-inverse}, and plugging that into \cref{eqn:ssfp-flow-velocity}, we get:
\begin{align}
    \intertext{\centering\underline{Conditional velocity field}}
    \begin{aligned}
    v_\xi^a(a, z, t) &= \dot{\xi}(t) - k\left( a - \xi(t) \right)+  \frac{\sigma_r \, (1 + kt)}{1 - (1 - \sigma_1)t}\left(z - t \xi(t)\right) \\
    v_\xi^z(a, z, t) &= \xi(t) + t \dot{\xi}(t) - \frac{1 - \sigma_1}{1 - (1 - \sigma_1)t} \left(z - t \xi(t)\right)
    \end{aligned}
    \label{eqn:ssfp-velocity-field}
\end{align}
Importantly, the evolution of $a$ and $z$ is inter-dependent \ie the sample $z_0$ determines the evolution of $a$.
Furthermore, the marginal probability distribution $p_\xi^a(a, t)$ can be deduced from the joint distribution in \cref{eqn:ssfp-joint-distribution} and is given by:
\begin{align}
    p_\xi(a \mid t) = \N\left(a \,\big\vert\, \ \xi(t) \,,\, \sigma_0^2 e^{-2kt} + \sigma_r^2 t^2 \right)
\end{align}
In other words, $q$ evolves in a Gaussian tube centered at the demonstration trajectory $\xi(t)$ with a standard deviation that varies from $\sigma_0$ at $t=0$ to $\sigma_1$ at $t=1$.
The fact that the marginal distribution lies close to the demonstration trajectories, from \cref{eq:averaged-marginal-distributions} ensures that the per-timestep marginal distributions over actions induced by the learned velocity field are close to training distribution.
However, this formulation allows us to select extremely small values of $\sigma_0$, essentially deterministically starting from the last generated action $a_\mathrm{prev}$.
The stochasticity injected by sampling $z_0 \in \N(0, I)$, as well as the correlated evolution of $a$ and $z$ ensures that we sample a diverse distribution of actions in $a$ starting from the same action $a_\mathrm{curr}$.
This phenomenon is illustrated via a 1-D toy example in \cref{fig:stochastic-sfp-conditional,fig:stochastic-sfp-marginal}, with details in captions.

\section{Action Horizon}
\label{app:action-horizon}

In \cref{fig:action-chunk-sizes}, we analyze the effect of action chunk size on the performance of streaming flow policy, under various benchmark environments: (1) \texttt{Robomimic:~Can}, (2) \texttt{Robomimic:~Square}, (3) \texttt{Push-T} with state input and (4) \texttt{Push-T} with image input.
The $x$-axis shows the chunk size in log scale.
The $y$-axis shows the relative decrease in performance compared to that of the best performing chunk size. All scores are less than or equal to zero, where higher is better.
In 3/4 environments, the performance peaks at chunk size 8, and 1/4 environments peak at chunk size 6.
The performance decreases as the chunk size deviates from the optimum.
Our results match with findings from \citet{chi2023diffusion}, suggesting that behavior cloning policies have a ``sweet spot'' in the chunk size of the action trajectories.
We recommend choosing a larger chunk size (\ie closer to open-loop execution) when the environment dynamics are deterministic and stable.
Smaller chunk sizes should be used in stochastic environments with high uncertainty, where the policy may benefit from a tighter feedback loop.

{
\begin{figure}[h!]
\centerline
{
  \includegraphics[trim=0 0 0 0,clip,width=0.55\textwidth]{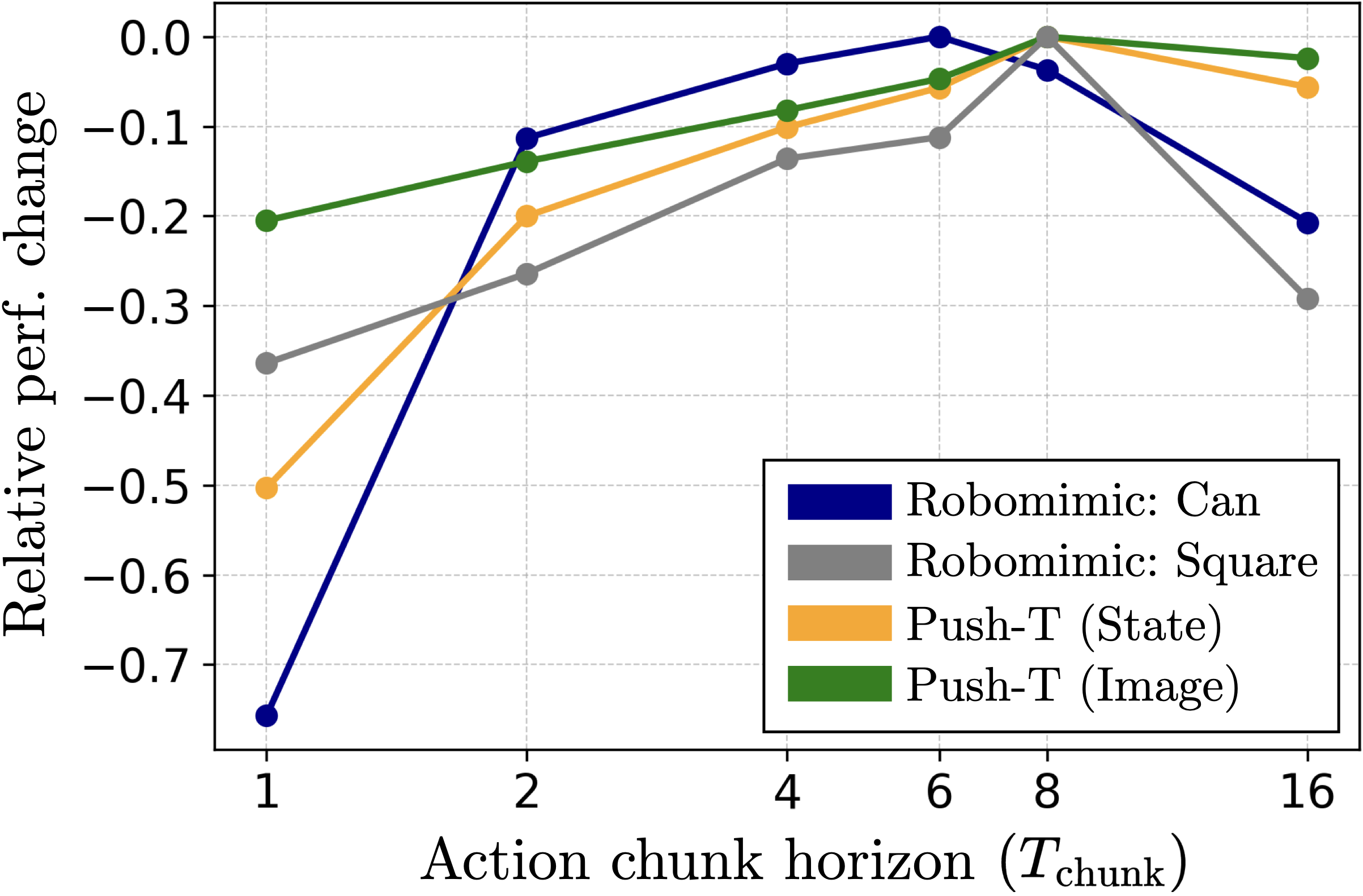}
}
\captionof{figure}
{
  \small
  Analysis of the effect of action chunk size on the performance of streaming flow policy, under various benchmark environments.
  $x$-axis shows the chunk size, in log scale.
  $y$-axis shows the relative decrease in performance compared to that of the best performing chunk size. All scores are less than or equal to zero, where higher is better.
  In 3/4 environments, the performance peaks at chunk size 8, and the other environment peaks at chunk size 6.
  The performance decreases as the chunk size increases or decreases from the optimum.
}
\label{fig:action-chunk-sizes}
\end{figure}
}
\section{Push-T experiments with image inputs and action imitation}
\label{appendix:more-pusht-results}

In \cref{table:results-pusht-appendix}, we perform experiments in the Push-T environment~\cite{chi2023diffusion,florence2022implicit} using images as observations, and imitating actions instead of states (see \cref{sec:experiments} for a discussion on state imitation vs. action imitation).
This was missing in \cref{table:results-pusht} of the main paper.

The conclusions from the table are essentially the same as in the main paper.
Streaming flow policy performs nearly as well as the best performing baseline \ie diffusion policy with 100 DDPM inference steps.
However, streaming flow policy is significantly faster than diffusion policy.
It is also faster than the remaining baselines, while also achieving a higher task success rate.

{
\newcommand{\emphasis}[1]{#1}
\newcommand{\hthickness}{1.3pt}
\newcommand{\vthickness}{1.3pt}
\newcommand{\NA}{\it \texttt{Not Applicable} \cellcolor{gray!20}}
\newcolumntype{I}{!{\vrule width \vthickness}}
\newcolumntype{|}{!{\vrule width 0.8pt}}
\newcolumntype{:}{!{\vrule width 0pt}}

\renewcommand{\arraystretch}{1.14}  

\begin{table*}[h!]

    \setlength\tabcolsep{6pt} 
    \small
    \centerline{
    \begin{tabular}{Ic|cIc|cI}
        \Xhline{\hthickness}
        \multicolumn{2}{IcI}{
            \multirow{3}{*}[-0.5ex]{
                \vspace*{5.5em}
                \includegraphics[width=0.065\textwidth]{figures/envs/pusht.png}
            }
        } & 
        \multicolumn{2}{cI}{Push-T with image input}\\
        \Xcline{3-4}{\hthickness}
        \multicolumn{2}{IcI}{} & 
        {\makecell{Action imitation\vphantom{$\Big|$}\\Avg/Max scores}} &
        Latency \\
        \cline{3-4}
        \multicolumn{2}{IcI}{} & 
        {$\uparrow$} & $\downarrow$ \\
            \Xhline{\hthickness}
            \rowcolor{\redcolor}
            1 & DP~\cite{chi2023diffusion}: 100 DDPM steps & \textbf{83.8\%} $/$ \textbf{87.0\%} & 127.2 ms \\
            \cline{1-4}
            \rowcolor{\redcolor}
            2 & DP~\cite{chi2023diffusion}: \ \ 10 DDIM steps & 80.8\% $/$ 85.5\% & 10.4 ms \\
            \cline{1-4}
            \rowcolor{\redcolor}
            3 & Flow matching policy~\cite{zhang2024affordance} & 67.9\% $/$ 69.3\% & 12.9 ms \\
            \cline{1-4}
            \rowcolor{\redcolor}
            4 & Streaming DP~\cite{hoeg2024streaming} & 80.5\% $/$ 83.9\% & 77.7 ms \\
            \Xhline{\hthickness}
            \rowcolor{\greencolor}
            5 & \textbf{SFP (Ours)} & 82.5\% $/$ \textbf{87.0\%} & \textbf{08.8 ms} \\
            \Xhline{\hthickness}
        \end{tabular}
    }
    \vspace{-0.4em}
    \caption{
        \small
        Imitation learning accuracy on the Push-T~\cite{chi2023diffusion} dataset with images as observation inputs, and imitating action trajectories.
        \greenbox{2pt}~Our method (in green) compared against \redbox{2pt}~baselines (in red). See text for details.
    }
    \label{table:results-pusht-appendix}
\end{table*}
}
\section{CoRL rebuttal responses}
In this section, we share our responses during the CoRL 2025 rebuttal process, to help clarify our work to reader who might have similar questions or concerns as our reviewers did.

\newcommand{\heading}[1]{\textbf{\ding{228} \textit{``#1"} }}

\heading{Streaming flow policy is a special case of diffusion$/$flow policy with prediction horizon and denoising steps equal to 1.}
We believe that there are important differences between the two. \textit{(i)} At training time, streaming flow policy with prediction horizon learns to forecast 16 future actions given an input observation history, whereas diffusion$/$flow policy with prediction horizon $T_p=1$ only learns to predict a single action. This forces streaming flow policy to ``implicitly learn a (task-relevant) dynamics model'' (Sec. 4.5 in \citet{chi2023diffusion}) which is not the case when using diffusion$/$flow policy with only. \textit{(ii)} At test time, diffusion$/$flow policy with runs \textit{closed-loop} i.e. the input observation history is updated after every single action. Whereas streaming flow policy with predicts future actions conditioned on the \textit{same observation history} and executes this chunk \textit{open-loop}. Executing action sequences open-loop is an important feature of action-chunking~\cite{chi2023diffusion,zhao2023learning}; open-loop action chunks are known to reduce distribution shift and improve imitation learning performance.

\heading{Noise dilemma: excessive noise in the first action reduces accuracy, while insufficient noisecompromises multimodality.}
Our approach attempts to get the best of both worlds by adding noise at training time but not at test time. \textit{(i)} Representing multimodality is more critical at training time to avoid learning the ``average" of multimodal demonstrations~\cite{chi2023diffusion}. Thus we add sufficient noise at train time to make it easy to learn multimodal distributions. \textit{(ii)} However, at test time, we make the learned policy deterministic by not adding any noise(setting $\sigma_0 = 0$) similar to~\cite{zhao2023learning}. This generates accurate, noise-free trajectories at test time. It is acceptable for the policy to be deterministic at test time and sample only one mode (there always exists an optimal deterministic policy).

\heading{Multimodality analysis has only been performed in the toy 1D environment.}
During the rebuttal period, we evaluated the multimodality of rollouts of streaming flow policy in the \mbox{Push-T} environment along the lines of \citet{chi2023diffusion}. We found that streaming flow policy samples multimodal trajectories in symmetric configurations \cref{fig:pusht-multimodality} (a, b) as reported by \citet{chi2023diffusion}.
Going a step further, we also analyzed streaming flow policy in \textit{asymmetric} configurations \cref{fig:pusht-multimodality} (c, d) where we also found that streaming flow policy samples are highly multimodal.

{
\begin{figure}[h!]
\centerline
{
  \includegraphics[trim=0 0 0 0,clip,width=0.9\textwidth]{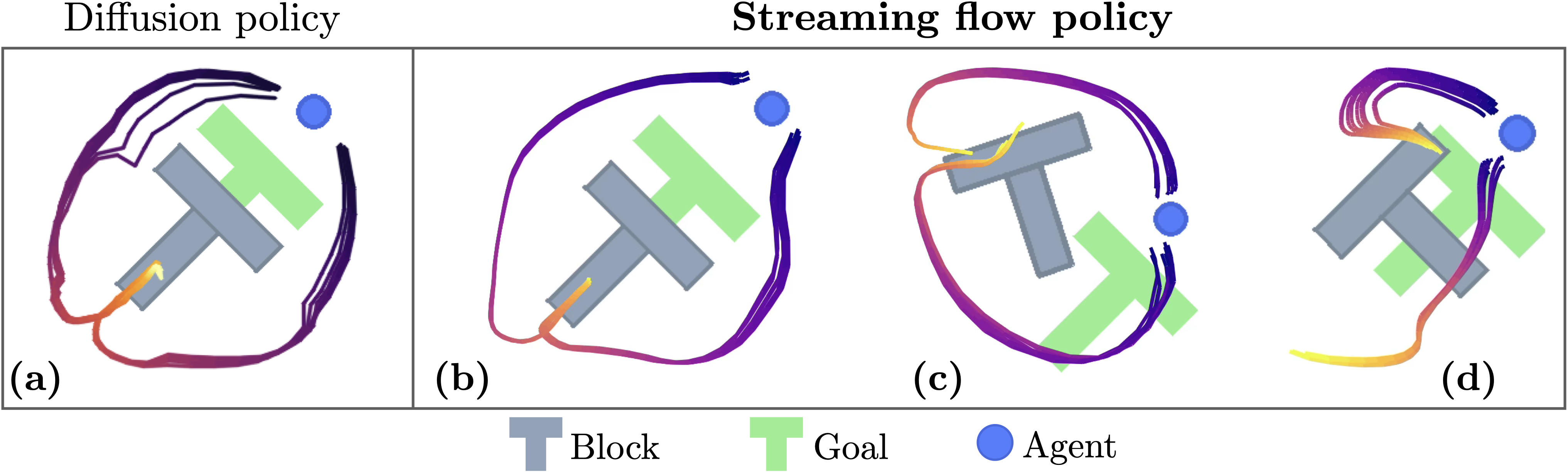}
}
\vspace*{-0.5em}
\captionof{figure}
{
  \small
  Streaming flow policy can sample multimodal action trajectories, similar to diffusion policy~\cite{chi2023diffusion} trained on the Push-T dataset.
  \textbf{(a)}
  \citet{chi2023diffusion} report that a diffusion policy trained on the Push-T dataset samples multimodal action trajectories starting from initial configurations that are symmetric with respect to block, goal pose and the agent (pusher).
  \textbf{(b)} We find the same to hold true for a streaming flow policy trained using the same neural network architecture on the Push-T dataset.
  \textbf{(c, d)} Going a step further, we visualize sampled action trajectories from \textit{asymmetric} configurations. Streaming flow policy generates multimodal trajectories, capturing diverse but valid behaviors to push the block towards the goal pose.
}
\label{fig:pusht-multimodality}
\end{figure}
}

\heading{Did you compare Streaming Flow Policy with feed-forward velocity field prediction?}
In the rebuttal period, we implemented this baseline and found that it performed significantly worse than streaming flow policy on \mbox{\texttt{Push-T-State}} (82.3\% vs. 95.1\%), \mbox{\texttt{Push-T-action}} (72.1\% vs. 91.7\%) and RoboMimic-Can (12.4\% vs. 95.6\%). We hypothesize that the poor performance is due to not sampling action noise during the training process. Thus,the baseline does not inherit guarantees from \citet{lipman2023flow} about matching the marginal training distributions.

\heading{Streaming flow policy has similar latency as diffusion$/$flow policy with 10 denoising steps}
While the latency is similar between streaming flow policy and few-step diffusion$/$flow policy, the behavior of the two policies is very different. Consider that both have a latency $L$ over $T_a$ actions in a chunk. diffusion$/$flow policy must wait for $L$ seconds for denoising to complete before the first action can be executed; then all $T_a$ actions can be executed in quick succession. streaming flow policy \textit{spreads out} latency evenly across all actions --- it can start executing the first action after $L/T_a$ seconds, and every pair of consecutive actions have the same gap. This \textit{results in much smoother and reactive motion compared to diffusion policy, as shown in our real-world videos on our \href{\website}{\textbf{project website}}}. Furthermore, we find that diffusion policy suffers a reduction in task performance when we decrease the number of denoising steps to match streaming flow policy's speed.

\heading{What are the implications of needing access to time derivatives of action trajectories during training?}
Unlike diffusion$/$flow policies, our framework requires the action space to be \textit{continuous} and action trajectories to be \textit{continuous} and \textit{differentiable} with respect to time. This means that certain conventional action spaces, such as binary $\{0, 1\}$ gripper open$/$close actions, cannot be modeled by streaming flow policy. However, most real-world physical quantities are intrinsically continuous and differentiable with respect to time. For example, we had to modify discrete $\{0, 1\}$ gripper open$/$close actions to a continuous gripper open fraction in $[0, 1]$; the continuous action is supported by both the real robot and simulation benchmarks. We computed time-derivatives by piecewise-linear interpolation of action sequences. In practice, we found that piecewise-linear derivatives worked well for all tasks without needing more sophisticated smoothing techniques.

\heading{How sensitive is streaming flow policy to hyper-parameters $k, \sigma_0, \Delta t$?}
We found that streaming flow policy is not sensitive to any of $k, \sigma_0$ or $\Delta t$. The only significant performance drop occurs when the stabilizing gain $k$ is set exactly \textit{equal} to zero, because \textit{stabilization mitigates distribution shift during inference}. However, we found that streaming flow policy is insensitive to a wide range of $k > 0$.

\heading{Why does the Streaming Diffusion Policy~\cite{hoeg2024streaming} baseline have high latency despite streaming?}
The reason streaming diffusion policy~\cite{hoeg2024streaming} is slower than expected is that streaming diffusion policy iteratively denoises every action chunk and not every action. For $N=100$ denoising steps, prediction horizon $T_p$ and action horizon $T_a$, the latency per action for streaming diffusion policy is $N/T_p$, whereas it is $N/T_a$ for diffusion policy. For the choice of hyper-parameters $T_p \approx 16, T_a \approx 8$, used in all three papers (diffusion policy~\cite{chi2023diffusion}, streaming diffusion policy~\cite{hoeg2024streaming} and ours), streaming diffusion policy is only twice as fast as diffusion policy. This can be corroborated by Fig. 4a in \citet{hoeg2024streaming}. However, streaming flow policy is an order of magnitude faster than diffusion policy with $N=100$. Streaming diffusion policy can be sped-up by increasing $T_p$ but only at the cost of reduced task performance, as shown in Fig. 4b in \citet{hoeg2024streaming}.

\end{appendices}


\end{document}